\documentclass[twocolumn,10pt]{article}

\usepackage[margin=1in]{geometry}
\usepackage[T1]{fontenc}
\usepackage[utf8]{inputenc}

\usepackage{amsmath,amssymb}

\usepackage{graphicx}
\usepackage{booktabs}
\usepackage{multirow}
\usepackage{float}
\usepackage{subcaption}   

\usepackage{multicol}

\usepackage[table,xcdraw]{xcolor}

\usepackage{arydshln}

\usepackage{listings}

\usepackage[hidelinks]{hyperref}

\lstdefinelanguage{Prompt}{
  morekeywords={Output, JSON, label, return},
  sensitive=false,
  morecomment=[l]{//},
  morestring=[b]",
}

\title{Intelligent Documentation in Medical Education: Can AI Replace Manual Case Logging?}

\author{
Nafiz Imtiaz Khan, MSc\thanks{Corresponding author: Nafiz Imtiaz Khan, Department of Computer Science, University of California, Davis, CA 95616, USA. Email: \texttt{nikhan@ucdavis.edu}} \\
Department of Computer Science, University of California, Davis, CA, USA
\and
Kylie Cleland, BSc \\
Department of Radiology, University of California, Davis, CA, USA
\and
Vladimir Filkov, PhD \\
Department of Computer Science, University of California, Davis, CA, USA
\and
Roger Eric Goldman, PhD \\
Department of Radiology, University of California, Davis, CA, USA
}

\date{} 

\begin{document}
\maketitle

\begin{abstract}
\textbf{Objective:}
This study investigates the feasibility of using large language models (LLMs) to automate procedural case log documentation in radiology training. We evaluate whether AI can replace manual logging, identify which procedure types are most challenging for extraction, and assess integration into clinical workflows.

\textbf{Materials and Methods:}
We retrospectively analyzed 414 curated radiology reports authored by nine interventional radiology residents between 2018 and 2024. A set of candidate models, including local (Qwen-2.5) and commercial (Claude-3.5), were tested under instruction and chain-of-thought prompting. Performance was measured by sensitivity, specificity, and F1-score, along with inference time and token efficiency to estimate operational cost.

\textbf{Results:}
Both local and commercial LLMs outperformed the standard benchmark. Qwen-2.5 achieved F1-scores of 86.66 with chain-of-thought prompting, while Claude-3.5-Haiku reached an F1-score of 86.89\%. Commercial inference delivered sub-2s latency and concise outputs, while local deployment traded speed for lower recurring cost. Automation could save over 35 hours of manual annotation per resident annually.

\textbf{Discussion:}
LLMs can provide a scalable and accurate solution for radiology case log documentation. Optimizing for procedure-specific challenges and ensuring seamless integration with existing systems will be essential. Future work should validate across larger, multi-institution datasets and explore additional prompting strategies.

\textbf{Conclusion:}
LLMs show promise for automating radiology case log documentation, potentially reducing resident clerical burden. However, this single-institution feasibility study underscores the need for broader validation across diverse institutions, assessment of real-world workflow integration, and safeguards against misclassification before clinical adoption.

\end{abstract}

\noindent\textbf{Corresponding Author:} Nafiz Imtiaz Khan (\texttt{nikhan@ucdavis.edu})


\section{Introduction}
Accurate and comprehensive clinical activity documentation is paramount for medical education and professional practice \cite{1_nygaard2015general}. Among the most critical artifacts in this process are physician case logs, which serve as structured records that detail the types and frequencies of procedures performed by individual clinicians. These logs are integral to multiple high-stakes functions. They provide evidence for board certification, serve as official documentation for credentialing and privileging within hospitals, and support quality assurance and performance evaluations \cite{case_log_2,ACGME_CaseLog,ACGME_DR_Minimums,ACGME_IR_Minimums}. During medical training, case logs are especially vital, enabling program directors to monitor resident progress, ensure exposure to a sufficient variety and volume of procedures, and identify areas needing remediation. Importantly, these logs also underpin institutional compliance with accreditation requirements, such as those mandated by the Accreditation Council for Graduate Medical Education (ACGME)\footnote{https://www.acgme.org/}.

However, the current reality of case log documentation in many specialties, particularly radiology, is that the process remains highly manual, fragmented, and inefficient. Radiology trainees are typically required to self-report the procedures they perform by entering data into spreadsheets or institutional logging systems \cite{vesselle2024development}. Each entry often includes detailed metadata, such as the type of imaging modality, the anatomical site, the nature of the procedure, and whether it was performed independently or under supervision. According to Cox et al. \cite{Cox2018}, general residents spend a median of 23.7 hours per week—nearly 38\% of their duty hours—engaged with the electronic health record (EHR). While this reflects overall documentation burden rather than case log entry specifically, it highlights how clerical tasks consume significant trainee time. Given that a radiology resident may perform well over 1600 (typically 1500 to 1800) procedures throughout their residency, the cumulative documentation burden is substantial. This manual process consumes valuable time that could otherwise be devoted to clinical learning or scholarly activity. In addition, manual effort introduces considerable potential for errors and inconsistencies in the documentation. Moreover, the absence of real-time feedback mechanisms limits the utility of these logs as tools for formative assessment.

In an era where natural language processing (NLP) and machine learning (ML) technologies—particularly large language models (LLMs)—have made significant advancements \cite{20_he2025survey,20_naveed2023comprehensive,20_wei2022emergent}, there exists a compelling opportunity to reimagine how procedural case logs are created \cite{bizzo2021artificial,hosny2018artificial,sacoransky2024chatgpt,babar2021evaluating,Casey2021,Xavier2022,Pereira2024Review,Kim2022CPT,NEJM_AI_2024}. Radiology reports, routinely authored by radiology residents and attendings, provide comprehensive narrative summaries of diagnostic imaging encounters and are inherently rich in procedural detail. These reports are routinely produced as part of standard clinical documentation and are stored in digital or semi-structured formats within Electronic Health Records (EHRs) \cite{EHR}, Radiology Information Systems (RIS) \cite{RIS}, or Picture Archiving and Communication Systems (PACS) \cite{PACS}. Using LLMs to automatically extract structured procedural data from these unstructured narratives could generate accurate procedural case logs with minimal human intervention. Such automation promises not only to alleviate administrative burden but also to enhance the accuracy, completeness, and timeliness of documentation. At the same time, prior experience with clinical AI tools demonstrates the risks of overclaiming performance without rigorous validation \cite{Wong2021,Lyons2023}, underscoring the importance of cautious evaluation in this domain.

While the theoretical potential of LLMs in this space is enticing, several important objectives motivate this study. First, we assess whether AI can effectively automate procedural case log documentation in radiology training, examining the feasibility and performance of using LLMs to convert narrative radiology reports into structured, loggable entries. Second, we investigate whether performance varies across procedure categories, identifying which categories are difficult to classify, and explored the linguistic or structural reasons behind these disparities. Finally, we examine the translational challenges of real-world deployment, including cost, and processing time.

In sum, this is the first work that explores the intersection of artificial intelligence and procedural documentation in medical education by systematically evaluating the feasibility of automation, the variation in accuracy across procedure categories, and the practical considerations for integration into clinical and educational workflows. Ultimately, we envision a future in which intelligent systems reduce administrative burden, improve documentation quality, and allow medical trainees to devote more attention to patient care and professional development.

\section{Methodology}

This retrospective study was approved by the institutional review board. The requirement of informed consent for inclusion in the study was waived. No protected health information is included in the reported results.

\subsection{Data Collection}
We retrospectively collected a dataset of raw radiology reports from a tertiary-care academic medical center, spanning the period from October 2018 through September 2024. These reports were authored by one of nine current or former Interventional Radiology (IR) residents. From this larger pool (36,659 reports across 301 unique exam codes, including 162 distinct procedural codes), we curated a subset of 414 reports corresponding to 39 representative IR procedures. For each procedural exam code, at least one corresponding radiology report was selected for manual labeling. If a code was associated with two reports, both were labeled. For codes with three or more associated reports, a random sample of three reports was selected for labeling. This ensured inclusion across the full range of required categories, but we acknowledge the small sample size per category as a limitation. 

\subsection{Data Annotation}
Following data collection, we conducted a structured annotation process to determine which unique radiology procedures were mentioned in each report. This task involved identifying the presence or absence of any of 39 predefined radiology procedures, which were based directly on the American Board of Radiology\footnote{https://www.theabr.org/} Case Log. The ABR case log was selected due to the diversity of procedure categories, which closely mirrors individual hospital case log requirements for privileges and is more granular than the ACGME logs. Procedures were organized into three high-level categories: Vascular Diagnosis (8 procedures), Vascular Intervention (15), and Non-Vascular Intervention (16). The complete list is shown in Table~\ref{table:procedure_info}.

Two independent annotators reviewed each report to assign binary labels for each of the 39 procedures: one interventional radiology trainee and one board-certified interventional radiologist with six years of post-graduate experience serving as an Assistant Professor at UC Davis Health. A label of 1 was assigned if a procedure was explicitly or implicitly mentioned in the report, and 0 otherwise. The default label was 0, and annotators marked 1 only when confident a procedure was present. To assess labeling consistency, we calculated inter-annotator agreement using Cohen’s Kappa, which accounts for chance agreement \cite{IAA_2,Cohen_3}. The observed $\kappa=0.896$ indicated strong agreement. Discrepancies were resolved by consensus, and the adjudicated labels served as the reference standard.

\subsection{Data Pre-processing}
We applied lightweight pre-processing to make the reports suitable for model inference. All text was lowercased to reduce case-sensitivity bias. Non-informative sections such as “attestation” and “addendum” were removed, as they typically contain administrative or author-identifying information irrelevant to procedural content. Also, non-ASCII characters were stripped to ensure tokenization compatibility. 


\subsection{LLM Selection}
Our goal was to evaluate a set of local and commercial LLMs on their ability to extract procedural data from radiology reports. Importantly, this was an \textit{inference-only} study: no model training or fine-tuning was performed. All models were applied directly in zero-shot mode to the dataset, so a train/validation/test split was not applicable.

\subsubsection{Selection Criteria}
\textbf{\textit{Privacy and Compliance.}} Given the sensitive nature of clinical data, model deployment was constrained to HIPAA-compliant environments. For commercial evaluation, we used the AWS Bedrock platform\footnote{https://aws.amazon.com/bedrock/}, which provides HIPAA-compliant infrastructure. For local evaluation, models were run on an institutional secure server with HIPAA-aligned governance.

\textbf{\textit{Model Type.}} We selected \textit{instruction-tuned} models, optimized for following prompts and returning structured responses~\cite{35_instruct}, since these are well-suited for extraction tasks.

\textbf{\textit{Hardware Constraints.}} Selection was also informed by available compute resources. Our local secure server had limited GPU memory, excluding extremely large models. Server specifications are listed in Table B.1 of Appendix B.

\textbf{\textit{Model Version and Size.}} When multiple versions existed, we chose the largest feasible variant within our hardware and compliance constraints. For example, LLaMA 3.3-70B was used locally, while the most performant model on AWS Bedrock at the time (Claude-3.5-Haiku, February 2025) was selected commercially.

\subsubsection{Selected Models}
Six models were evaluated: five open-source/local and one commercial. These included \texttt{Mixtral-8x7B}, \texttt{LLaMA 3.3-70B}, \texttt{Qwen 2.5-72B}, \texttt{MedLLaMA-2-7B}, \texttt{LLaMA-3-Med42-70B}, and \texttt{Claude-3.5-Haiku}. Model details are provided in the Appendix E. No additional fine-tuning was performed; all models were evaluated as released by their developers.

\subsection{Benchmark Definition (Crosswalk)}
For comparison, we leveraged the “crosswalk” benchmark \cite{crosswalk}, which is a rule-based mapping between exam codes or CPT codes and predefined procedure categories. For example, CPT code \texttt{75625} corresponds to “Lower Extremity Angiography.” This approach, while widely used and straightforward, is limited by coding variability and inability to capture multi-procedure reports. As such, it offers high specificity but reduced sensitivity, and serves as a baseline against which LLM outputs can be compared. 

\subsection{Prompt Engineering}
Two prompting strategies were tested with each model: (i) \textit{Instruction Prompting}, where the task was explicitly stated and the model was expected to return the output directly \cite{IP_1_11,11_IP_2}, and (ii) \textit{Chain-of-Thought (CoT)} prompting, where the model was guided to reason step-by-step before generating the final answer \cite{CoT_1_10,CoT_2_10}. CoT was implemented via a standardized template enforcing stepwise reasoning and a consistent output structure. No few-shot examples or fine-tuned prompts were introduced, to maintain consistency across models.

\subsection{Pipeline Implementation}
We developed a modular pipeline in Python to support reproducible evaluation. The pipeline managed preprocessing, model querying (local via Ollama \cite{ollama}, cloud via AWS Bedrock), and logging of results. Outputs were parsed into structured JSON and saved with metadata (model, prompt type, report ID, timestamp) for traceability. This ensured consistent handling across all models and prompt strategies.

\subsection{Model Evaluation}
We evaluated model outputs against the adjudicated ground truth labels. Metrics included True Positive (TP), True Negative (TN), False Positive (FP), False Negative (FN), Sensitivity, Specificity, and F1-score \cite{22_rahman2020reliable,23_shenoy2021covxr}. We also tracked inference time and token counts to estimate computational and financial costs. Performance was summarized across all 39 procedure categories, with per-category error analysis conducted to identify systematically difficult categories. All evaluations were conducted under consistent experimental conditions, following best practices such as STARD-AI for transparent reporting in medical AI \cite{ESR2023}.

\section{Results}

\subsection{Automating Procedural Case Log Documentation}
We evaluated model performance using a test set of 414 radiology reports annotated by independent reviewers. The inter-annotator agreement, measured using Cohen’s Kappa, was 0.896—indicating strong consistency. Disagreements were resolved through consensus to establish the final ground truth used for evaluation.

Table~\ref{table:LLMEvaluation} presents a detailed comparison between the Crosswalk benchmark, the top-performing local model (\textit{Qwen-2.5:72B}), and the top-performing commercial model (\textit{Claude-3.5-Haiku}) across two prompting strategies—Instruction Prompting (IP) and Chain-of-Thought (CoT)—as well as across three procedural modality categories. The performance of other local models is provided in Table C.2 in the Appendix C.

The Crosswalk benchmark, which relies solely on structured metadata from radiology reports, displays excellent specificity across all categories (overall specificity: 99.40\%). This confirms its conservative design, prioritizing false positive minimization. However, its sensitivity remains low (65.46\%), indicating substantial under-detection of documented procedures, particularly those mentioned only in free-text. This trade-off is most pronounced in the \textit{Vascular Intervention} category, where sensitivity drops to 59.02\%, suggesting that the benchmark fails to capture many interventional procedures lacking structured documentation. While \textit{Vascular Diagnosis} and \textit{Non-Vascular Intervention} fare slightly better (sensitivity: 63.84\% and 75.88\%, respectively), they still reflect the benchmark's inherent limitation in sensitivity.

In contrast, the local model \textit{Qwen-2.5:72B} demonstrates marked improvements in both sensitivity and overall performance when augmented with prompting strategies. Under Instruction Prompting, Qwen achieves a sensitivity of 94.19\% and specificity of 98.17\%, yielding a notable gain of +28.73\% points in sensitivity over the Crosswalk baseline, without substantially compromising specificity. This results in an F1-score of 80.08—a lift over the benchmark. CoT prompting further improves the balance between sensitivity and specificity, with Qwen attaining a higher F1-score (86.66) among all local configurations. This is particularly driven by CoT’s effectiveness in reducing false positives in the \textit{Non-Vascular Intervention} category (FP: 38 vs. 122 under IP), highlighting its ability to support more nuanced reasoning in complex textual contexts.

The commercial model \textit{Claude-3.5-Haiku} similarly benefits from both prompting paradigms. With Instruction Prompting, it achieves a sensitivity of 91.87\%, specificity of 96.79\%, and F1-score of 69.64. However, CoT prompting enhances performance further: the F1-score rises to 86.89; the highest across all models and configurations—accompanied by an improved specificity of 99.29\%. These results suggest that Claude’s CoT-based reasoning is particularly best-suited for the intricacies of radiology language. On a per-modality basis, Claude demonstrates outstanding sensitivity in \textit{Vascular Diagnosis} (95.98\% under IP) and achieves the most consistent performance across all categories under CoT.

A broader trend across all evaluated models is that \textit{Vascular Diagnosis} consistently yields the highest sensitivity and F1-scores. This is likely due to the use of standardized terms and clear procedural language in the diagnostic procedural categories, which facilitates model understanding. On the other hand, \textit{Vascular Intervention} remains the most challenging category, especially for the Crosswalk benchmark, but also for models operating under Instruction Prompting. The improved results from CoT prompting—particularly for Qwen (F1: 82.88) and Claude (F1: 85.55)—highlight the benefit of step-by-step reasoning in interpreting diverse and often implicit procedural descriptions typical of interventional reports.

Overall, these findings underscore the limitations of metadata-only benchmarks for procedure detection and illustrate the potential of LLMs, especially when equipped with effective prompting strategies, to more accurately and comprehensively extract procedural information from unstructured clinical text.

\begin{table*}[]
\centering
\caption{Performance of the Crosswalk Benchmark Compared to the Top-Performing Local and Commercial Language Model}
\label{table:LLMEvaluation}
\def\arraystretch{1.25}
\resizebox{\textwidth}{!}
{
\begin{tabular}{lllllllllll}
\toprule
\textbf{Model Type} & \textbf{Model-Name} & \textbf{\begin{tabular}[c]{@{}l@{}}Prompting \\ Method\end{tabular}} & \textbf{Modality} & \textbf{TP} & \textbf{TN} & \textbf{FP} & \textbf{FN} & \textbf{Sensitivity (\%)} & \textbf{Specificity (\%)} & \textbf{F1-Score (\%)} \\
\midrule
\multirow{4}{*}{Benchmark} & \multirow{4}{*}{Cross-Walk} & \multirow{4}{*}{NA} & All & 451 & 15364 & 93 & 238 & 65.46 & 99.40 & \textbf{73.15} \\
 &  &  & VascularDiagonsis & 143 & 3065 & 23 & 81 & 63.84 & 99.26 & 73.33 \\
 &  &  & VascularIntervention & 157 & 5906 & 38 & 109 & 59.02 & 99.36 & 68.11 \\
 &  &  & NonVascularIntervention & 151 & 6393 & 32 & 48 & 75.88 & 99.50 & 79.06 \\ \hline
\multirow{8}{*}{Local} & \multirow{8}{*}{Qwen-2.5:72B} & \multirow{4}{*}{IP} & All & 649 & 15174 & 283 & 40 & 94.19 & 98.17 & \textbf{80.08} \\
 &  &  & VascularDiagnosis & 219 & 3068 & 20 & 5 & 97.77 & 99.35 & 94.60 \\
 &  &  & VascularIntervention & 247 & 5803 & 141 & 19 & 92.86 & 97.63 & 75.54 \\
 &  &  & NonVascularIntervention & 183 & 6303 & 122 & 16 & 91.96 & 98.10 & 72.62 \\ \cdashline{3-11}
 &  & \multirow{4}{*}{CoT} & All & 627 & 15326 & 131 & 62 & 91.00 & 99.15 & \textbf{86.66} \\
 &  &  & VascularDiagnosis & 214 & 3071 & 17 & 10 & 95.54 & 99.45 & 94.07 \\
 &  &  & VascularIntervention & 242 & 5868 & 76 & 24 & 90.98 & 98.72 & 82.88 \\
 &  &  & NonVascularIntervention & 171 & 6387 & 38 & 28 & 85.93 & 99.41 & 83.82 \\ \hline
\multirow{8}{*}{Commercial} & \multirow{8}{*}{Claude-3.5-Haiku} & \multirow{4}{*}{IP} & All & 633 & 14961 & 496 & 56 & 91.87 & 96.79 & \textbf{69.64} \\
 &  &  & VascularDiagnosis & 215 & 3067 & 21 & 9 & 95.98 & 99.32 & 93.48 \\
 &  &  & VascularIntervention & 230 & 5737 & 207 & 36 & 86.47 & 96.52 & 65.43 \\
 &  &  & NonVascularIntervention & 188 & 6157 & 268 & 11 & 94.47 & 95.83 & 57.41 \\ \cdashline{3-11}
 &  & \multirow{4}{*}{CoT} & All & 613 & 15348 & 109 & 76 & 88.97 & 99.29 & \textbf{86.89} \\
 &  &  & VascularDiagnosis & 210 & 3069 & 19 & 14 & 93.75 & 99.38 & 92.71 \\
 &  &  & VascularIntervention & 228 & 5905 & 39 & 38 & 85.71 & 99.34 & 85.55 \\
 &  &  & NonVascularIntervention & 175 & 6374 & 51 & 24 & 87.94 & 99.21 & 82.35 \\
\bottomrule
\end{tabular}}
\end{table*}

\subsection{Variation Across Procedure Types}

Table D.4 in appendix D  reports the number of false positives (FP) and false negatives (FN) committed by three algorithms—the Crosswalk benchmark, the top-performing commercial model with CoT prompting (\textit{Claude-3.5-Haiku}), and the top-performing local model with CoT prompting (\textit{Qwen-2.5:72B})—across 39 radiological procedures. This fine-grained error analysis provides insight into procedure-level performance and helps assess whether certain procedures are consistently harder to detect.

Compared to the Crosswalk benchmark (652 FNs, 244 FPs), both LLMs achieve significant reductions in false negatives: \textit{Claude-3.5-Haiku} reduces FNs to 76, and \textit{Qwen-2.5:72B} to 62. However, this gain in sensitivity comes at the cost of increased false positives: Claude reports 109 FPs, while Qwen reports 131—still lower than Crosswalk's FP count.

\textbf{Poorly Classified Procedures.} Certain procedures remain challenging for all models, with high error rates suggesting inherent complexity or ambiguity. Notably, \textbf{Procedure 23} (Other, Vascular) and \textbf{Procedure 39} (Other, NonVascular) exhibit the highest cumulative errors across all three systems. For example, Qwen reports 45 false positives on Procedure 23 and both models struggle with Procedure 39, where errors remain concentrated due to overlapping semantics and subtle contextual clues.

\textbf{Well-Classified Procedures.} Several procedures such as Procedure 2, Procedure 11, Procedure 18, and Procedure 28 are consistently well-classified across all models. These likely benefit from clearer textual cues or isolated contexts. Procedures like Procedure 12, Procedure 16, and Procedure 30 show modest improvements in both FP and FN rates with the LLMs compared to Crosswalk.

\textbf{Model-Specific Patterns.} \textit{Claude-3.5-Haiku} maintains a relatively balanced error profile with moderate FPs and low FNs. In contrast, \textit{Qwen-2.5:72B} demonstrates aggressive prediction behavior—especially in Procedure 23—resulting in more FPs but fewer FNs overall. This behavior may be attributed to the model's higher confidence threshold or broader generalization strategy.




\begin{table*}[]
\centering
\caption{Per Procedure Identiciation Cost of Running the Language Model in Relation to Inference Time and Generated Tokens}
\label{table:Cost-Token-Inference}
\def\arraystretch{1.25}
\resizebox{\textwidth}{!}
{
\begin{tabular}{lllllll|llll}
\toprule
\multicolumn{1}{c}{\multirow{2}{*}{\textbf{Model Name}}} & \multicolumn{1}{c}{\multirow{2}{*}{\textbf{\begin{tabular}[c]{@{}c@{}}Model \\ Type\end{tabular}}}} & \multicolumn{1}{c}{\multirow{2}{*}{\textbf{\begin{tabular}[c]{@{}c@{}}Prompting\\ Method\end{tabular}}}} & \multicolumn{4}{c}{\textbf{Inference Time (Second)}} & \multicolumn{4}{c}{\textbf{Generated Tokens (Count)}} \\  
\multicolumn{1}{c}{} & \multicolumn{1}{c}{} & \multicolumn{1}{c}{} & \multicolumn{1}{c}{\textbf{Max}} & \multicolumn{1}{c}{\textbf{Min}} & \multicolumn{1}{c}{\textbf{Mean}} & \multicolumn{1}{c}{\textbf{Std Dev}} & \multicolumn{1}{c}{\textbf{Max}} & \multicolumn{1}{c}{\textbf{Min}} & \multicolumn{1}{c}{\textbf{Mean}} & \multicolumn{1}{c}{\textbf{Std Dev}} \\

\midrule

\multirow{2}{*}{Qwen2.5:72B} & \multirow{2}{*}{Local} & IP  & 20.00 & 4.00 & 9.70 & 2.15 & 127 & 27 & 63.01 & 14.39 \\
                             &                         & CoT & 48.00 & 9.00 & 13.47 & 3.23 & 111 & 26 & 53.98 & 12.27 \\
\hline
\multirow{2}{*}{Claude-3.5-Haiku} & \multirow{2}{*}{Commercial} & IP  & 5.00 & 1.00 & 1.97 & 0.70 & 67 & 13 & 40.26 & 9.77 \\
                                  &                              & CoT & 15.00 & 3.00 & 6.97 & 2.15 & 80 & 10 & 29.79 & 9.60 \\

\hline
\end{tabular}}
\end{table*}

\begin{table*}[]
\centering
\caption{Comparative Cost of Running the Most Performant Models in Commercial (Claude-3.5-Haiku via AWS Bedrock) and Local (Qwen2.5:72B) Environments}
\label{table:Cost-Comparison}
\begin{tabular}{llrrrr}
\toprule
\textbf{Model} & \shortstack{\textbf{Prompting}\\\textbf{Method}} & \textbf{Category} & \shortstack{\textbf{Input}\\\textbf{/ Time}} & \textbf{Output Tokens} & \textbf{Cost (USD)} \\
\midrule
\multirow{4}{*}{\shortstack[l]{Claude-3.5-Haiku\\(Commercial, AWS Bedrock)}} 
& IP  & Procedure & 837.41 tokens & 40.05 tokens  & \$0.0008 \\
&     & Report    & 32658.96 tokens & 1561.79 tokens & \$0.0324 \\
& CoT & Procedure & 1232.77 tokens & 30.84 tokens  & \$0.0011 \\
&     & Report   & 48077.96 tokens & 1202.84 tokens & \$0.0433 \\
\midrule
\multirow{4}{*}{\shortstack[l]{Qwen2.5:72B\\(Local Server)}} 
& IP  & Procedure & 9.70 s & \textemdash{} & \$0.00094 \\
&     & Report    & 378.30 s & \textemdash{} & \$0.03670 \\
& CoT & Procedure & 13.47 s & \textemdash{} & \$0.00131 \\
&     & Report    & 525.33 s & \textemdash{} & \$0.05096 \\
\bottomrule
\end{tabular}
\end{table*}


\begin{table*}[]
\centering
\caption{Categorization of procedures into predefined modality groups: \texttt{Vascular Diagnosis}, Vascular Intervention, and Non-Vascular Intervention}
\label{table:procedure_info}
{
\begin{tabular} {p{1.5cm}p{10cm}p{2cm}}
\toprule
\textbf{Count} & \textbf{Procedure Description} & \textbf{Modality} \\

\midrule
1 & Computed Tomography Angiography (CTA) & \multirow{8}{*}{\shortstack{Vascular\\Diagnosis}} \\
2 & Magnetic Resonance Angiography (MRA) &  \\
3 & Noninvasive vascular lab (duplex, color flow, PVRs, etc.)  &  \\
4 & Cardiac Imaging &  \\
5 & Arteriography (all: peripheral, renal mesenteric, carotid, etc.)    &  \\
6 & Venography (all) &  \\
7 & Dialysis access evaluations   &  \\
8 & Carotid artery imaging   &  \\
 \hline
9 & Venous access (all: tunneled, nontunneled, ports) & \multirow{15}{*}{\shortstack{Vascular\\Intervention}} \\
10 & IVC filter placement, retrieval  &  \\
11 & Venous ablation (varicose veins)  &  \\
12 & Dialysis access intervention  &  \\
13 & TIPS \& TIPS evaluation/revision   &  \\
14 & Angioplasty/stents/covered stents: arterial (peripheral renal, mesenteric)   &  \\
15 & Angioplasty/stents/covered stents: venous (all)   &  \\
16 & Carotid stenting   &  \\
17 & Thrombolytic therapy (all), thrombectomy   &  \\
18 & Aortic endografting (thoracic and/or abdominal)   &  \\
19 & Embolization, emergency (trauma, GI bleed, bronchial bleed, other) &  \\
20 & Embolization, elective (uterine fibroids, PAVMs, peripheral AVMs, varicoceles, etc.)  &  \\
21 & Chemoembolization (TACE)  &  \\
22 & Radioembolization (selective internal radiotherapy)  &  \\
23 & Other, Vascular &  \\
\hline
24 & Biopsy & \multirow{16}{*}{\shortstack{Non Vascular\\Intervention}} \\
25 & Abscess drainage \& tube checks   &  \\
26 & Paracentesis, thoracentesis   &  \\
27 & Chest tube placement   &  \\
28 & Pleurodesis   &  \\
29 & PTC, biliary drainage, biliary stents; tube checks   &  \\
30 & Nephrostomy, nephroureterostomy; tube checks   &  \\
31 & Gastrostomy, gastrojejunostomy; tube checks  &  \\
32 & Cholecystostomy; tube checks   &  \\
33 & Aspiration, drainage, sclerosis (cyst, lymphocele); tube checks   &  \\
34 & Stents, miscellaneous nonvascular (esophageal, tracheobronchial, duodenal, colonic)   &  \\
35 & Transplant interventions, miscellaneous  &  \\
36 & Tumor ablation (RFA, laser, microwave, cryo, ethanol, other)   &  \\
37 & Pain management   &  \\
38 & Fallopian tube recanalization   &  \\
39 & Other, Nonvascular &  \\
\bottomrule

\end{tabular}}
\end{table*}

\subsection{Practical Deployment Considerations}

Table~\ref{table:Cost-Token-Inference} presents descriptive statistics—including maximum, minimum, mean, and standard deviation—for inference time and generated token count for the \texttt{Claude-3.5 Haiku} and \texttt{Qwen2.5:72B} models across the two prompting strategies: IP and CoT. The reported descriptive statistics reflect per-procedure identification. Therefore, multiplying these values by 39 yields the corresponding per-report estimates. 

Additionally, the cost metrics (inference time and generated token) of the additional local models are presented in Table C.3 in appendix C.



\textbf{Inference Time.}
Inference Time/Latency is a key operational constraint in any real-world deployment. Among the evaluated configurations, the commercial model \textit{Claude-3.5-Haiku} using Instruction Prompting (IP) demonstrates the fastest average per-procedure inference time—1.97 seconds—with a minimum of just 1 second and maximum of 5 seconds. This near-instant responsiveness positions Claude as a highly practical solution for interactive and high-volume use cases. In contrast, the local model \textit{Qwen-2.5:70B}, while effective in terms of accuracy, demonstrates a higher computational burden. Under Chain-of-Thought (CoT) prompting, it reaches a mean inference time of 13.47 seconds per procedure, with outliers extending to 48 seconds. This latency may constrain scalability, particularly in batch-processing or real-time scenarios, unless sufficient parallelization or hardware resources are available.

\textbf{Token Efficiency and Output Volume.}
Token generation affects not only the cost of the model, especially in commercial settings, but also the interpretability and processing overhead. \textit{Claude-3.5-Haiku} again leads in efficiency, with CoT prompting producing an average of 29.79 tokens per procedure, and IP generating slightly longer responses at 40.26 tokens. By contrast, Qwen’s IP configuration yields significantly longer outputs (mean: 63.01 tokens), suggesting more verbose justifications. Although longer responses can provide more rationale, they also increase downstream token-related costs and may slow user consumption of the output. Thus, from a human-in-the-loop perspective, Claude's concise formatting is more aligned with practical review and integration.

Table 3 details the estimated cost of running the \texttt{Qwen2.5:72B} model in a local deployment environment, providing both per procedure and per report costs in USD for the two strategies—IP and CoT. These cost estimates are based on the model’s average inference time, which serves as a practical proxy for computational resource utilization in local settings. Specifically, the cost of running the local model is computed by multiplying the per-procedure inference time by the estimated per-second operational expense of the local server. This estimate incorporates hardware depreciation, electricity consumption, and general infrastructure overhead. The detailed cost breakdown for operating the local system is outlined below.

\textbf{Cost Calculation for Local Model.}
We performed inference on an NVIDIA A100 GPU with 24GB VRAM equipped server. To quantify its operational cost, we considered three main components: hardware depreciation, power consumption, and recurring overhead (e.g., maintenance, physical space, and administrative costs).
The A100 GPU with 24GB VRAM has an upfront cost of approximately \$8,000~USD \cite{cost_ref_castonguay2025cost,cost_ref_li2021tesla}. Commercial entities typically amortize such hardware over three years but we adopt a more conservative five-year depreciation schedule to reflect longer usage cycles common in academic environments. This results in an hourly depreciation cost of roughly \$0.23.

The estimated power draw of the server—including GPU, CPU, memory, storage, and cooling infrastructure—is approximately 400 watts \cite{mytton2025datacenters}. At an average electricity rate of \$0.19 per kilowatt-hour \cite{electricity_cost} (US average), this corresponds to \$0.06 in energy costs per hour. To account for auxiliary operational expenses, we add a 20\% overhead margin to the base cost, contributing an additional \$0.06 per hour.

Summing these components, the total estimated cost of operating the server is \$0.35 per hour. When expressed on a per-second basis, this translates to an operational cost of approximately \textbf{\$0.000097}, or \textbf{0.01 cents per second}. This estimate serves as a realistic baseline for evaluating the cost-efficiency of locally hosted LLM inference in academic and research settings.

Table 3 presents the per-procedure and per-report cost of deploying the \texttt{Claude-3.5-Haiku} model in a commercial environment via AWS Bedrock. Unlike local deployments, commercial usage is priced based on token consumption—typically charged per input and output token. Thus, for the commercial model, costs were calculated based on the number of input and generated tokens per procedure under each prompting strategy.

\textbf{Cost Calculation for the Commercial Model.}
Unlike local deployments, commercial language models such as \texttt{Claude-3.5-Haiku} are billed based on token usage, with separate rates for input and output tokens. According to the official AWS Bedrock pricing documentation \cite{aws2025bedrock}, the cost is \$0.0008 per 1,000 input tokens and \$0.004 per 1,000 output tokens.

Therefore, for a given inference task—such as identifying whether a procedure was performed—if the input prompt contains 1,000 tokens and the model generates a 1,000-token response, the total cost for that inference would be \$0.0008+\$0.004=\$0.0048.

\textbf{Comparative Cost of Local vs. Commercial Deployment.}
For the local model, see Table 3, with a per-second operating cost of \$0.000097 (derived above), the estimated cost per 9.7s procedure is \$0.00094 using IP and \$0.00131 using CoT. Extrapolated to full reports containing an average of 39 procedures, the per-report costs are \$0.03670 (IP) and \$0.05096 (CoT)—affordable in many academic and clinical research settings.

On the commercial side, Table 3 shows IP prompts cost \$0.0008 per procedure and \$0.0324 per report, while CoT prompts cost slightly more (\$0.0011 per procedure; \$0.0433 per report), due to increased input token length. These values indicate that Claude's commercial deployment is competitive with local deployment of Qwen, particularly when factoring in maintenance-free scalability, security compliance, and minimal hardware investment.

\textbf{Comparison with Manual Logging Burden.}
Radiology residents are responsible for documenting upwards of 1,600 reports during the time of residency, with each report averaging 39 distinct procedural entries. Assuming just 2 seconds per manual entry, this amounts to:

\begin{align*}
\text{Total manual annotation time} 
&= 1{,}600 \times 39 \times 2 \text{ seconds} \\
&= 124{,}800 \text{ seconds} \\
&\approx 35 \text{ hours}.
\end{align*}

This represents almost one full-time workweek spent on clerical documentation per resident—time that could be redirected toward patient care, educational development, or research. Furthermore, manual annotation is susceptible to human error, variability in documentation diligence, and inconsistent understanding of procedural terminology—all of which degrade data quality in high-stakes applications like credentialing and competency tracking.

\section{Discussion}
\subsection{Limitations \& Future Work}
This study has several limitations that open avenues for future research.

First, the analysis was performed on a relatively small dataset comprising 414 radiology reports. This limited sample size, driven primarily by data availability and labeling time constraints, may affect the generalizability of the findings in broader clinical contexts. Future work should aim to validate these results using a larger and more diverse dataset spanning multiple institutions, imaging modalities, and patient demographics.

Second, our evaluation focused on a single commercially available, most up-to-date model on the Bedrock platform—\texttt{Claude-3.5-Haiku} (as of February 15, 2025). Although it demonstrated strong performance across prompting strategies, the commercial LLM landscape is evolving rapidly, with emerging models offering varying trade-offs in speed, cost, and reasoning capabilities. Future studies should conduct comparative evaluations across a wider range of commercial LLMs from different HIPAA-compliant platforms to identify the most suitable models for specific clinical use cases and infrastructure constraints.

Third, we limited our experimentation to two prompting strategies: IP and CoT. Although these are among the most commonly adopted prompting paradigms, recent advances have introduced a variety of alternatives—including few-shot prompting, self-consistency prompting, and tool-augmented prompting; that may further enhance model performance, interpretability, or robustness. Incorporating and benchmarking these strategies could provide deeper insights into optimizing LLM-driven clinical automation.

Fourth, while this study focused on procedural extraction from radiology reports, the broader clinical documentation landscape includes many other entities, such as diagnoses, findings, measurements, and recommendations, that were not addressed here. Expanding the scope of LLM applications to include these additional dimensions could contribute to more comprehensive automation and structured documentation in healthcare workflows.

Finally, the evaluation was limited to a single institution with a specific reporting style and workflow. Radiology documentation conventions differ across hospitals, specialties, and even individual clinicians. As such, the reported performance may not generalize to other settings. Multi-institutional validation across varied EHR and PACS systems is essential before clinical deployment.

Future research should therefore pursue multi-site collaborations, larger annotated corpora, and expanded prompting paradigms. Equally important is assessing usability and trust among residents, faculty, and credentialing bodies, as automation bias or over-reliance on AI outputs could compromise oversight. Integrating LLM-based tools in a way that supports, rather than replaces, human review will be critical. By addressing these gaps, the field can move toward safe, equitable, and scalable automation of clinical documentation in medical education.

\subsection{Conclusion}

This study evaluated the use of large language models to automate procedural case log documentation in radiology training. Both open-weight (Qwen-2.5:72B) and commercial (Claude-3.5-Haiku) models outperformed a metadata-only benchmark, reliably identifying common procedures while struggling with categories marked by vague or overlapping language. Operational analyses showed that local and commercial deployments offer broadly comparable costs, with trade-offs between speed, scalability, and data privacy. These findings suggest that LLMs are a feasible complement to current documentation workflows, offering the potential to reduce clerical burden, though further validation across institutions, procedures, and prompting strategies is required to ensure safe integration into training and clinical practice.

\section*{Code \& Data Availability}
The project codebase and its documentation are available on a dedicated website, accessible here: \url{https://nafiz43.github.io/PCL-Fetcher/}. Due to confidentiality, the data has not been made available.

\section*{Funding}
No funding was received for this work.

\bibliographystyle{vancouver}
\bibliography{REFERENCE}


\begin{thebibliography}{99}
\bibitem{1_nygaard2015general} Nygaard, Rachel M and Daly, Samuel R and Van Camp, Joan M. \textit{General surgery resident case logs: do they accurately reflect resident experience?}. Journal of surgical education. Elsevier. 2015.
\bibitem{case_log_2} Cadish, Lauren A and Fung, Vicki and Lane, Felicia L and Campbell, Eric G. \textit{Surgical case logging habits and attitudes: a multispecialty survey of residents}. Journal of surgical education. Elsevier. 2016.
\bibitem{ACGME_CaseLog} {Accreditation Council for Graduate Medical Education}. \textit{ACGME Case Log System}. 2025. \url{https://www.acgme.org/program-directors-and-coordinators/case-log-system}. Accessed: 2025-09-17.
\bibitem{ACGME_DR_Minimums} {Accreditation Council for Graduate Medical Education}. \textit{Diagnostic Radiology Case Log Minimums}. 2025. \url{https://www.acgme.org/specialties/diagnostic-radiology/program-requirements-and-faqs-and-applications/}. Accessed: 2025-09-17.
\bibitem{ACGME_IR_Minimums} {Accreditation Council for Graduate Medical Education}. \textit{Interventional Radiology--Integrated Case Log Minimums}. 2025. \url{https://www.acgme.org/specialties/interventional-radiology/program-requirements-and-faqs-and-applications/}. Accessed: 2025-09-17.
\bibitem{vesselle2024development} Vesselle, Hubert and Chiramal, Justy Antony and Hawes, Stephen E and Schulze, Eric and Nguyen, Tham and Ndumia, Rose and Vinayak, Sudhir. \textit{Development of an online authentic radiology viewing and reporting platform to test the skills of radiology trainees in Low-and Middle-Income Countries}. BMC Medical Education. Springer. 2024.
\bibitem{Cox2018} Cox, Christopher and others. \textit{Documentation Workload and Its Impact on Resident Workflow in Graduate Medical Education}. Journal of Graduate Medical Education. 2018. doi: 10.4300/JGME-D-17-00745.1.
\bibitem{20_he2025survey} He, Kai and Mao, Rui and Lin, Qika and Ruan, Yucheng and Lan, Xiang and Feng, Mengling and Cambria, Erik. \textit{A survey of large language models for healthcare: from data, technology, and applications to accountability and ethics}. Information Fusion. Elsevier. 2025.
\bibitem{20_naveed2023comprehensive} Naveed, Humza and Khan, Asad Ullah and Qiu, Shi and Saqib, Muhammad and Anwar, Saeed and Usman, Muhammad and Akhtar, Naveed and Barnes, Nick and Mian, Ajmal. \textit{A comprehensive overview of large language models}. arXiv preprint arXiv:2307.06435. 2023.
\bibitem{20_wei2022emergent} Wei, Jason and Tay, Yi and Bommasani, Rishi and Raffel, Colin and Zoph, Barret and Borgeaud, Sebastian and Yogatama, Dani and Bosma, Maarten and Zhou, Denny and Metzler, Donald and others. \textit{Emergent abilities of large language models}. arXiv preprint arXiv:2206.07682. 2022.
\bibitem{bizzo2021artificial} Bizzo, Bernardo C and Almeida, Renata R and Alkasab, Tarik K. \textit{Artificial intelligence enabling radiology reporting}. Radiologic Clinics. Elsevier. 2021.
\bibitem{hosny2018artificial} Hosny, Ahmed and Parmar, Chintan and Quackenbush, John and Schwartz, Lawrence H and Aerts, Hugo JWL. \textit{Artificial intelligence in radiology}. Nature Reviews Cancer. Nature Publishing Group UK London. 2018.
\bibitem{sacoransky2024chatgpt} Sacoransky, Ethan and Kwan, Benjamin YM and Soboleski, Donald. \textit{ChatGPT and assistive AI in structured radiology reporting: A systematic review}. Current Problems in Diagnostic Radiology. Elsevier. 2024.
\bibitem{babar2021evaluating} Babar, Zaheer and van Laarhoven, Twan and Zanzotto, Fabio Massimo and Marchiori, Elena. \textit{Evaluating diagnostic content of AI-generated radiology reports of chest X-rays}. Artificial Intelligence in Medicine. Elsevier. 2021.
\bibitem{Casey2021} Casey, Adrian and Banerjee, Ishita and Langer, Steve and Lu, Zhiyong. \textit{Natural language processing for clinical radiology reports: a systematic review}. Journal of the American College of Radiology. 2021. doi: 10.1016/j.jacr.2020.11.014.
\bibitem{Xavier2022} Xavier, Jean and Franco, Alvaro and Costa, Fernando. \textit{Extracting structured information from radiology reports using natural language processing: a scoping review}. Insights into Imaging. 2022. doi: 10.1186/s13244-022-01162-9.
\bibitem{Pereira2024Review} Pereira, Ricardo and others. \textit{Applications of natural language processing in radiology: a narrative review}. European Radiology. 2024. doi: 10.1007/s00330-024-10874-7.
\bibitem{Kim2022CPT} Kim, Jin and Patel, Sneha and Wang, Xin and others. \textit{Automated CPT code prediction from clinical notes using deep learning}. JMIR Medical Informatics. 2022. doi: 10.2196/35678.
\bibitem{NEJM_AI_2024} Nguyen, Thao and Chen, David and Lee, James. \textit{Artificial intelligence for coding in healthcare: promise and pitfalls}. NEJM AI. 2024. doi: 10.1056/AI0056.
\bibitem{EHR} Seymour, Tom and Frantsvog, Dean and Graeber, Tod. \textit{Electronic health records (EHR)}. American Journal of Health Sciences. The Clute Institute. 2012.
\bibitem{RIS} Rebelo, Lu{\'\i}s and Fidalgo, Filipe and Oliveira, {\^A}ngela. \textit{Radiology Information System (RIS)}. 2022 17th Iberian Conference on Information Systems and Technologies (CISTI). IEEE. 2022.
\bibitem{PACS} Choplin, Robert H and Boehme 2nd, JM and Maynard, C Douglas. \textit{Picture archiving and communication systems: an overview.}. Radiographics. 1992.
\bibitem{Wong2021} Wong, Alicia and Otles, Erkin and Donnelly, Janelle P. and others. \textit{External validation of a widely implemented proprietary sepsis prediction model in hospitalized patients}. JAMA Internal Medicine. 2021. doi: 10.1001/jamainternmed.2021.2626.
\bibitem{Lyons2023} Lyons, Patrick G. and others. \textit{Systematic bias and poor performance of a commercial sepsis prediction algorithm}. Critical Care Medicine. 2023. doi: 10.1097/CCM.0000000000005810.
\bibitem{IAA_2} Artstein, Ron. \textit{Inter-annotator agreement}. Handbook of linguistic annotation. Springer. 2017.
\bibitem{Cohen_3} Kv{\aa}lseth, Tarald O. \textit{Note on Cohen's kappa}. Psychological reports. SAGE Publications Sage CA: Los Angeles, CA. 1989.
\bibitem{35_instruct} Zhang, Shengyu and Dong, Linfeng and Li, Xiaoya and Zhang, Sen and Sun, Xiaofei and Wang, Shuhe and Li, Jiwei and Hu, Runyi and Zhang, Tianwei and Wu, Fei and others. \textit{Instruction tuning for large language models: A survey}. arXiv preprint arXiv:2308.10792. 2023.
\bibitem{crosswalk} Wang, Kenneth C and Patel, Jigar B and Vyas, Bimal and Toland, Michael and Collins, Beverly and Vreeman, Daniel J and Abhyankar, Swapna and Siegel, Eliot L and Rubin, Daniel L and Langlotz, Curtis P. \textit{Use of radiology procedure codes in health care: the need for standardization and structure}. Radiographics. Radiological Society of North America. 2017.
\bibitem{IP_1_11} Zhang, Shengyu and Dong, Linfeng and Li, Xiaoya and Zhang, Sen and Sun, Xiaofei and Wang, Shuhe and Li, Jiwei and Hu, Runyi and Zhang, Tianwei and Wu, Fei and others. \textit{Instruction tuning for large language models: A survey}. arXiv preprint arXiv:2308.10792. 2023.
\bibitem{11_IP_2} Marvin, Ggaliwango and Hellen, Nakayiza and Jjingo, Daudi and Nakatumba-Nabende, Joyce. \textit{Prompt engineering in large language models}. International conference on data intelligence and cognitive informatics. Springer. 2023.
\bibitem{CoT_1_10} Wei, Jason and Wang, Xuezhi and Schuurmans, Dale and Bosma, Maarten and Xia, Fei and Chi, Ed and Le, Quoc V and Zhou, Denny and others. \textit{Chain-of-thought prompting elicits reasoning in large language models}. Advances in neural information processing systems. 2022.
\bibitem{CoT_2_10} Yu, Zihan and He, Liang and Wu, Zhen and Dai, Xinyu and Chen, Jiajun. \textit{Towards better chain-of-thought prompting strategies: A survey}. arXiv preprint arXiv:2310.04959. 2023.
\bibitem{ollama} Marcondes, Francisco S and Gala, Adelino and Magalh{\~a}es, Renata and Perez de Britto, Fernando and Dur{\~a}es, Dalila and Novais, Paulo. \textit{Using Ollama}. Natural Language Analytics with Generative Large-Language Models: A Practical Approach with Ollama and Open-Source LLMs. Springer. 2025.
\bibitem{22_rahman2020reliable} Rahman, Tawsifur and Khandakar, Amith and Kadir, Muhammad Abdul and Islam, Khandaker Rejaul and Islam, Khandakar F and Mazhar, Rashid and Hamid, Tahir and Islam, Mohammad Tariqul and Kashem, Saad and Mahbub, Zaid Bin and others. \textit{Reliable tuberculosis detection using chest X-ray with deep learning, segmentation and visualization}. Ieee Access. IEEE. 2020.
\bibitem{23_shenoy2021covxr} Shenoy, Vishal and Malik, Sachin. \textit{CovXR: automated detection of COVID-19 pneumonia in chest X-rays through machine learning}. 2021 IEEE Symposium Series on Computational Intelligence (SSCI). IEEE. 2021.
\bibitem{ESR2023} {European Society of Radiology}. \textit{Structured reporting in radiology: update from the European Society of Radiology (ESR)}. Insights into Imaging. 2023. doi: 10.1186/s13244-023-01457-5.
\bibitem{cost_ref_castonguay2025cost} Danny Castonguay. \textit{Cost of Inference}. 2025. \url{https://blog.dannycastonguay.com/Cost-of-Inference/}. Accessed: 2025-05-18.
\bibitem{cost_ref_li2021tesla} Chuan Li. \textit{Tesla A100 Server Total Cost of Ownership Analysis}. 2021. \url{https://lambda.ai/blog/tesla-a100-server-total-cost-of-ownership}.
\bibitem{mytton2025datacenters} David Mytton. \textit{How much energy do data centers use?}. 2025. \url{https://davidmytton.blog/how-much-energy-do-data-centers-use/}. Accessed: 2025-05-18.
\bibitem{electricity_cost} {EnergySage}. \textit{How much does electricity cost in 2025?}. 2025. \url{https://www.energysage.com/local-data/electricity-cost/}. Accessed: 2025-05-18.
\bibitem{aws2025bedrock} {Amazon Web Services}. \textit{Amazon Bedrock Pricing}. 2025. \url{https://aws.amazon.com/bedrock/pricing/}. Accessed: 2025-05-18.
\bibitem{mixtral_hf_6} Mistral AI. \textit{Mixtral-8x7B-v0.1}. 2023. \url{https://huggingface.co/mistralai/Mixtral-8x7B-v0.1}.
\bibitem{qwen25_hf_5} Qwen Team. \textit{Qwen2.5: A Party of Foundation Models}. 2024. \url{https://qwenlm.github.io/blog/qwen2.5/}.
\bibitem{medllama2_hf} Siraj Raval. \textit{MedLLaMA2 7B}. Hugging Face. 2023. \url{https://huggingface.co/llSourcell/medllama2_7b}. Accessed: 2025-04-13.
\bibitem{med42v2_paper_7} Cl{\'e}ment Christophe and Praveen K Kanithi and Tathagata Raha and Shadab Khan and Marco AF Pimentel. \textit{Med42-v2: A Suite of Clinical LLMs}. 2024. \url{https://arxiv.org/abs/2408.06142}.
\bibitem{med42v2_hf_7} M42 Health. \textit{Llama3-Med42-70B}. 2024. \url{https://huggingface.co/m42-health/Llama3-Med42-70B}.
\bibitem{anthropic2024claude35haiku_8} Anthropic. \textit{Claude 3.5 Haiku}. 2024. \url{https://www.anthropic.com/claude/haiku}.
\end{thebibliography}

\onecolumn

\appendix

\section{Prompt Templates}
\begin{lstlisting}[caption={Prompt Template for Chain-of-Thought prompting, in which the \textless procedure\_specific\_question\textgreater\ placeholder is dynamically replaced with a specific question tailored to the target radiology procedure}, label={lst:cot_prompt}]
I will provide you with a radiology report, followed by a question about whether a specific radiology study or procedure was performed.

<procedure\_specific\_question>

### **Strict Output Format**
Your response **must** be a **valid JSON object** with the following keys:
```json
{
  "reason_for_the_label": "A concise explanation justifying the classification.",
  "label": 1 or 0
}
\end{lstlisting}

\begin{lstlisting}[caption={Prompt Template for Instruction Prompting}, label={lst:ip_prompt}]

I will provide you with a radiology report, followed by several questions about it. Your task is to determine whether a specific radiology study or procedure was performed. Please follow these strict formatting guidelines for your response:

Output must be in valid JSON format with the following keys:

{
  "reason_for_the_label": "A string explaining the reasoning behind the classification.",
  "label": 1 or 0
}

Labeling criteria:
Return 1 if the radiology study or procedure was explicitly mentioned as performed.
Return 0 if the study or procedure was not performed, not documented, or uncertain in the report.
Do not include any additional text or explanations outside the JSON response.
Ensure strict adherence to this format for every response
\end{lstlisting}




\section{Server Configuration}
We conducted all the experiments for our study on a dedicated local server with a well-defined configuration to ensure consistency and reliability in our results. The configuration of the server is presented in table \ref{table:ServerConfig}.
\begin{table*}[h]
\centering
\caption{Local server configuration}
\label{table:ServerConfig}
{
\begin{tabular}{ll}
\hline
\textbf{Component} & \textbf{Configuration} \\
\hline
Primary Memory (RAM) & 64 GB \\
\hline
Secondary Memory (Hard Disk) & 512 GB SSD \\
\hline
Processor & 
\begin{tabular}{@{}l@{}}
\textit{Name}: Intel(R) Xeon(R) W-2135 CPU @ 3.70GHz \\
\textit{Architecture}: x86\_64 \\
\textit{CPU(s)}: 12 \\
\textit{Thread(s) per core}: 2 \\
\textit{Core(s) per socket}: 6 \\
\textit{Max Clock Speed}: 4500.00 MHz \\
\textit{Min Clock Speed}: 1200.00 MHz \\
\textit{PCI Express Lanes}: 48 \\
\textit{Integrated Graphics}: None \\
\textit{TDP (Thermal Design Power)}: 140 W \\
\end{tabular} \\
\hline
Cache & 
\begin{tabular}{@{}l@{}}
\textit{L1d cache}: 192 KiB (6 instances) \\
\textit{L1i cache}: 192 KiB (6 instances) \\
\textit{L2 cache}: 6 MiB (6 instances) \\
\textit{L3 cache}: 8.3 MiB (1 instance) \\
\end{tabular} \\
\hline
GPU & 
\begin{tabular}{@{}l@{}}
\textit{Name}: A100 (PCIe) \\
\textit{Memory}: 24 GB HBM2 \\
\textit{Memory Interface Width}: 3840 bits \\
\textit{Memory Bandwidth}: 1555 GB/s \\
\textit{Base Clock Speed}: 1095 MHz \\
\textit{Boost Clock Speed}: 1410 MHz \\
\textit{Memory Clock Speed}: 1215 MHz \\
\textit{CUDA Cores}: 6912 \\
\textit{Tensor Cores}: 432 (3rd Gen) \\
\textit{Thermal Design Power (TDP)}: 250 W \\
\end{tabular} \\
\hline
\end{tabular}

}
\end{table*}

\section{Performance \& Cost of the Additional Local Models}
The performance and cost metrics of the additional local models are presented in Table 2 and Table 3, respectively. Table 2 highlights their effectiveness across different tasks and prompting strategies, while Table 3 provides a breakdown of the associated computational costs

\begin{table*}[]
\centering
\caption{Performance of the Crosswalk Benchmark Compared to the Top-Performing Local and Commercial Language Model}
\label{table:LLMEvaluation_Appendix}
\def\arraystretch{1.25}
\resizebox{\textwidth}{!}
{
\begin{tabular}{llllllllll}
\hline

\textbf{Model-Name} & \textbf{\begin{tabular}[c]{@{}l@{}}Prompting \\ Method\end{tabular}} & \textbf{Modality} & \textbf{TP} & \textbf{TN} & \textbf{FP} & \textbf{FN} & \textbf{Sensitivity (\%)} & \textbf{Specificity (\%)} & \textbf{F1-Score (\%)} \\
\hline
\midrule
\multirow{8}{*}{Llama3-Med42:70B} & IP & All & 662 & 14714 & 743 & 27 & 96.08 & 95.19 & 63.23 \\
 & IP & VascularDiagnosis & 215 & 3039 & 49 & 9 & 95.98 & 98.41 & 88.11 \\
 & IP & VascularIntervention & 257 & 5559 & 385 & 9 & 96.62 & 93.52 & 56.61 \\
 & IP & NonVascularIntervention & 190 & 6116 & 309 & 9 & 95.48 & 95.19 & 54.45 \\ \cdashline{2-10}
 & CoT & All & 645 & 15037 & 420 & 44 & 93.61 & 97.28 & 73.54 \\
 & CoT & VascularDiagnosis & 211 & 3042 & 46 & 13 & 94.20 & 98.51 & 87.73 \\
 & CoT & VascularIntervention & 250 & 5816 & 128 & 16 & 93.98 & 97.85 & 77.64 \\
 & CoT & NonVascularIntervention & 184 & 6179 & 246 & 15 & 92.46 & 96.17 & 58.50 \\ \hline
\multirow{8}{*}{Llama3.3:70B} & IP & All & 662 & 14826 & 631 & 27 & 96.08 & 95.92 & 66.80 \\
 & IP & VascularDiagnosis & 215 & 3048 & 40 & 9 & 95.98 & 98.70 & 89.77 \\
 & IP & VascularIntervention & 256 & 5647 & 297 & 10 & 96.24 & 95.00 & 62.51 \\
 & IP & NonVascularIntervention & 191 & 6131 & 294 & 8 & 95.98 & 95.42 & 55.85 \\ \cdashline{2-10}
 & CoT & All & 628 & 15191 & 266 & 61 & 91.15 & 98.28 & 79.35 \\
 & CoT & VascularDiagnosis & 200 & 3052 & 36 & 24 & 89.29 & 98.83 & 86.96 \\
 & CoT & VascularIntervention & 249 & 5834 & 110 & 17 & 93.61 & 98.15 & 79.68 \\
 & CoT & NonVascularIntervention & 179 & 6305 & 120 & 20 & 89.95 & 98.13 & 71.89 \\ \hline
\multirow{8}{*}{Medllama2:7B} & IP & All & 321 & 8909 & 6548 & 368 & 46.59 & 57.64 & 8.49 \\
 & IP & VascularDiagnosis & 126 & 1959 & 1129 & 98 & 56.25 & 63.44 & 17.04 \\
 & IP & VascularIntervention & 107 & 3372 & 2572 & 159 & 40.23 & 56.73 & 7.26 \\
 & IP & NonVascularIntervention & 88 & 3578 & 2847 & 111 & 44.22 & 55.69 & 5.62 \\ \cdashline{2-10}
 & CoT & All & 605 & 3771 & 11686 & 84 & 87.81 & 24.40 & 9.32 \\
 & CoT & VascularDiagnosis & 206 & 885 & 2203 & 18 & 91.96 & 28.66 & 15.65 \\
 & CoT & VascularIntervention & 245 & 412 & 5532 & 21 & 92.11 & 6.93 & 8.11 \\
 & CoT & NonVascularIntervention & 154 & 2474 & 3951 & 45 & 77.39 & 38.51 & 7.15 \\ \hline
\multirow{8}{*}{Mixtral:8x7B} & IP & All & 623 & 14728 & 729 & 66 & 90.42 & 95.28 & 61.05 \\
 & IP & VascularDiagnosis & 194 & 3050 & 38 & 30 & 86.61 & 98.77 & 85.09 \\
 & IP & VascularIntervention & 243 & 5608 & 336 & 23 & 91.35 & 94.35 & 57.52 \\
 & IP & NonVascularIntervention & 186 & 6070 & 355 & 13 & 93.47 & 94.47 & 50.27 \\ \cdashline{2-10}
 & CoT & All & 601 & 14444 & 1013 & 88 & 87.23 & 93.45 & 52.20 \\
 & CoT & VascularDiagnosis & 199 & 2940 & 148 & 25 & 88.84 & 95.21 & 69.70 \\
 & CoT & VascularIntervention & 234 & 5500 & 444 & 32 & 87.97 & 92.53 & 49.57 \\
 & CoT & NonVascularIntervention & 168 & 6004 & 421 & 31 & 84.42 & 93.45 & 42.64 \\
\bottomrule

\end{tabular}}
\end{table*}

\begin{table*}[]
\centering
\caption{Cost of Running the Language Model in Relation to Inference Time and Generated Tokens for the Rest of the Local Models}
\label{table:LLMEvaluation}
\def\arraystretch{1.25}
\resizebox{\textwidth}{!}
{
\begin{tabular}{llllll|lllll}
\toprule
\multicolumn{1}{c}{\multirow{2}{*}{\textbf{Model Name}}} & 
\multicolumn{1}{c}{\multirow{2}{*}{\textbf{\begin{tabular}[c]{@{}c@{}}Prompting\\ Method\end{tabular}}}} & 
\multicolumn{4}{c}{\textbf{Inference Time (Second)}} & 
\multicolumn{4}{c}{\textbf{Generated Tokens (Count)}} \\
\cdashline{3-10}
\multicolumn{1}{c}{} & 
\multicolumn{1}{c}{} & 
\textbf{Max} & \textbf{Min} & \textbf{Mean} & \textbf{Std Dev} & 
\textbf{Max} & \textbf{Min} & \textbf{Mean} & \textbf{Std Dev} \\

\midrule
\multirow{2}{*}{Llama3-Med42:70B} 
& IP  & 13.00  & 3.00   & 6.53  & 1.65  & 147    & 14   & 53.81  & 17.92 \\
& CoT & 18.00  & 7.00   & 10.44 & 1.80  & 152    & 34   & 70.65  & 16.61 \\ \hline
\multirow{2}{*}{Llama3:70B} 
& IP  & 330.00 & 2.00   & 57.32 & 21.66 & 433    & 1    & 71.03  & 29.69 \\
& CoT & 59.00  & 11.00  & 28.52 & 4.67  & 84     & 12   & 29.10  & 10.28 \\ \hline
\multirow{2}{*}{Medllama2:7B} 
& IP  & 1497.00 & 0.00  & 2.97  & 47.93 & 66420  & 13   & 135.31 & 2125.40 \\
& CoT & 1027.00 & 0.00  & 3.83  & 56.79 & 74564  & 5    & 244.15 & 4004.62 \\ \hline
\multirow{2}{*}{Mixtral:8x7B} 
& IP  & 32.00   & 0.00  & 0.98  & 1.17  & 116    & 1    & 30.20  & 17.07 \\
& CoT & 9.00    & 1.00  & 2.21  & 0.77  & 124    & 14   & 41.83  & 16.65 \\
\bottomrule

\hline
\end{tabular}
}
\end{table*}

\section{Model Error Count}
\begin{table*}[]
\centering
\caption{Classification error count (False Postive and False Negative) of the models on each of the procedures}
\label{table:error_count}
{
\begin{tabular}{l|ll|ll|ll}
\toprule
\multicolumn{1}{c}{\multirow{2}{*}{\textbf{\begin{tabular}[c]{@{}c@{}}Procedure \\ Number\end{tabular}}}} & \multicolumn{2}{c}{\textbf{Crosswalk}} & \multicolumn{2}{c}{\textbf{Claude-3.5-Haiku}} & \multicolumn{2}{c}{\textbf{Qwen-2.5:72B}} \\
\multicolumn{1}{c}{} & \multicolumn{1}{c}{\textbf{FP Count}} & \multicolumn{1}{c}{\textbf{FN Count}} & \multicolumn{1}{c}{\textbf{FP Count}} & \multicolumn{1}{c}{\textbf{FN Count}} & \multicolumn{1}{c}{\textbf{FP Count}} & \multicolumn{1}{c}{\textbf{FN Count}} \\

\midrule
1 & 1 & 21 & 8 & 0 & 10 & 0 \\
2 & 0 & 0 & 1 & 0 & 0 & 0 \\
3 & 0 & 9 & 5 & 0 & 3 & 0 \\
4 & 10 & 1 & 0 & 0 & 0 & 0 \\
5 & 29 & 77 & 1 & 2 & 1 & 3 \\
6 & 22 & 57 & 2 & 10 & 1 & 6 \\
7 & 0 & 20 & 1 & 2 & 1 & 0 \\
8 & 0 & 39 & 1 & 0 & 1 & 1 \\ \hline
9 & 0 & 13 & 2 & 4 & 2 & 6 \\
10 & 1 & 8 & 1 & 0 & 0 & 0 \\
11 & 0 & 0 & 3 & 0 & 2 & 0 \\
12 & 0 & 8 & 1 & 1 & 0 & 1 \\
13 & 6 & 3 & 0 & 1 & 0 & 1 \\
14 & 0 & 12 & 4 & 0 & 3 & 0 \\
15 & 0 & 74 & 3 & 2 & 5 & 1 \\
16 & 0 & 3 & 1 & 0 & 1 & 0 \\
17 & 15 & 36 & 2 & 0 & 2 & 0 \\
18 & 0 & 0 & 0 & 0 & 0 & 0 \\
19 & 30 & 62 & 3 & 14 & 1 & 9 \\
20 & 39 & 56 & 7 & 4 & 15 & 2 \\
21 & 0 & 3 & 1 & 1 & 0 & 1 \\
22 & 3 & 0 & 0 & 2 & 0 & 2 \\
23 & 6 & 24 & 11 & 9 & 45 & 1 \\ \hline
24 & 29 & 2 & 3 & 1 & 1 & 1 \\
25 & 14 & 14 & 5 & 1 & 1 & 3 \\
26 & 3 & 13 & 4 & 0 & 6 & 0 \\
27 & 0 & 4 & 0 & 0 & 0 & 0 \\
28 & 0 & 0 & 1 & 0 & 0 & 0 \\
29 & 7 & 8 & 1 & 1 & 0 & 1 \\
30 & 0 & 5 & 1 & 1 & 0 & 1 \\
31 & 0 & 3 & 5 & 0 & 3 & 0 \\
32 & 5 & 14 & 1 & 0 & 0 & 0 \\
33 & 4 & 4 & 2 & 1 & 1 & 1 \\
34 & 0 & 3 & 7 & 1 & 14 & 1 \\
35 & 0 & 9 & 6 & 0 & 5 & 0 \\
36 & 0 & 0 & 0 & 0 & 0 & 0 \\
37 & 0 & 6 & 0 & 5 & 0 & 6 \\
38 & 0 & 0 & 0 & 0 & 0 & 0 \\
39 & 20 & 41 & 15 & 13 & 7 & 14 \\ \hline
\textbf{Total} & \textbf{244} & \textbf{652} & \textbf{109} & \textbf{76} & \textbf{131} & \textbf{62} \\

\bottomrule

\end{tabular}}
\end{table*}

\newpage

\section{Description of the selected Models}
\texttt{Mixtral-8x7B}. Mixtral-8x7B \footnote{https://huggingface.co/mistralai/Mixtral-8x7B-v0.1} is a mixture-of-experts model released by Mistral AI. The 8x7B version activates 2 out of 8 expert pathways per token, which offers a balance between model expressiveness and computational efficiency. It has shown competitive performance across a range of multilingual and general-purpose NLP tasks.


\texttt{LLaMA 3.3-70B}. LLaMA 3 is the latest generation of Meta's foundational language models. The 70B version demonstrates significant gains in general understanding, reasoning, and instruction-following \footnote{https://huggingface.co/meta-llama/Llama-3.3-70B-Instruct}. It is particularly effective for downstream fine-tuning and multi-turn prompt applications.

\texttt{Qwen 2.5-72B}. Qwen 2.5 \footnote{https://qwenlm.github.io/blog/qwen2.5/} is a high-performance transformer model developed by Alibaba. The 72B version is instruction-tuned and optimized for accuracy in information extraction tasks. It supports a wide multilingual range and is designed for competitive performance in both zero- and few-shot settings.


\texttt{MedLLaMA-2-7B}. MedLLaMA-2-7B is a biomedical adaptation of Meta's LLaMA architecture, fine-tuned on large-scale medical corpora. The 7B parameter variant focuses on healthcare and clinical domains, making it suitable for domain-specific tasks such as radiology procedural report interpretation.


\texttt{LLaMA-3-Med42-70B}. LLaMA-3-Med42-70B \footnote{https://huggingface.co/m42-health/Llama3-Med42-70B} is a domain-specialized version of LLaMA 3.3-70B, fine-tuned on a curated medical dataset named Med42 \cite{med42v2_paper_7}. It is intended for complex clinical reasoning, procedural understanding, and structured medical data generation.


\texttt{Claude-3.5-Haiku}. Claude 3.5 Haiku \footnote{https://www.anthropic.com/claude/haiku} is a lightweight commercial LLM developed by Anthropic. Despite its smaller size, it exhibits strong instruction-following and chain-of-thought reasoning capabilities. It is optimized for fast response times while maintaining high accuracy in structured generation tasks.


\section{LLM Prompts}
This section outlines the various prompting techniques employed in the study.

\subsection{Instruction Prompting}
\texttt{\normalsize Prompt for Question  1:}
\begin{lstlisting}[]
Please answer 1 if the report explicitly states that a CTA (CT Angiography) or an intra-operative cone-beam CT with contrast was performed. If a CTA or intra-operative cone-beam CT with contrast was not performed or not mentioned, please answer 0.
\end{lstlisting}

\texttt{\normalsize Prompt for Question  2:}
\begin{lstlisting}[]
Please answer 1 if the report specifies an MRA or MR Angiography was performed. Otherwise, please answer 0
\end{lstlisting}

\texttt{\normalsize Prompt for Question  3:}
\begin{lstlisting}[]
Please answer 1 if the report specifies a noninvasive vascular lab study was performed. Examples include a duplex, color flow, or Pulse Volume Recording (PVR). Otherwise, please answer 0
\end{lstlisting}

\texttt{\normalsize Prompt for Question  4:}
\begin{lstlisting}[]
Please answer 1 if the report specifies a cardiac imaging study was performed. Examples include a CT coronary angiogram or nuclear medicine cardiac stress test. Otherwise, please answer 0
\end{lstlisting}

\texttt{\normalsize Prompt for Question  5:}
\begin{lstlisting}[]
Please answer 1 if a dedicated arteriography was performed. Arteriography is defined as imaging of arteries using fluoroscopy and contrast specifically to evaluate arterial anatomy or pathology. The contrast must be specifically administered into an artery. Arteriography does not include venography. Answr 0 if arteriography was not performed or not documented
\end{lstlisting}

\texttt{\normalsize Prompt for Question  6:}
\begin{lstlisting}[]
Please answer 1 if a dedicated venography was performed . Venorgaphy is defined as imaging of veins using fluoroscopy and contrast specifically to evaluate venous anatomy or pathology. The contrast must be specifically administered into a vein. Venography does not include arteriography. Answer 0 if no dedicated venography was performed
\end{lstlisting}

\texttt{\normalsize Prompt for Question  7:}
\begin{lstlisting}[]
Please answer 1 if a dialysis access evaluation was performed. These evaluations include ultrasound studies, fistulagrams, and graft evaluations. Answer 0 if a dialysis access evaluation was not performed or not documented
\end{lstlisting}

\texttt{\normalsize Prompt for Question  8:}
\begin{lstlisting}[]
Please answer 1 if carotid artery imaging was performed. Answer 0 if carotid artery imaging was not performed or not documented
\end{lstlisting}

\texttt{\normalsize Prompt for Question  9:}
\begin{lstlisting}[]
Please answer 1 if a central venous catheter procedure was performed. This includes placement, removal, and revision of central venous catheters, ports, and PICCs. Answer 0 if a central venous catheter procedure was not performed or not documented
\end{lstlisting}

\texttt{\normalsize Prompt for Question  10:}
\begin{lstlisting}[]
Please answer 1 if an IVC filter placement or removal procedure was performed. Answer 0 if an IVC filter procedure was not performed or not documented
\end{lstlisting}

\texttt{\normalsize Prompt for Question  11:}
\begin{lstlisting}[]
Please answer 1 if a venous ablation procedure was performed. Venous ablations are performed for varicose veins,. Answer 0 if venous ablation was not performed or not documented
\end{lstlisting}

\texttt{\normalsize Prompt for Question  12:}
\begin{lstlisting}[]
Please answer 1 if a dialysis access intervention was performed. These interventions inlcude fistulagrams, graft evaluations, and declot procedures specifically and only on arteriovenous access. Answer 0 if a dialysis access intervention was not performed or not documented
\end{lstlisting}

\texttt{\normalsize Prompt for Question  13:}
\begin{lstlisting}[]
Please answer 1 if any intervention involving a portosystemic shunt was performed. Typical shunts include Transjugular Intrahepatic Portosystemic Shunt (TIPS), Direct intrahepatic portocaval shunt (DIPS) and transjugular transcaval intrahepatic portosystemic shunt (TTIPS). Interventions inlcude shunt placement, evaluation, or revision. Answer 0 if a portosystemic shunt intervention was not performed or not documented
\end{lstlisting}

\texttt{\normalsize Prompt for Question  14:}
\begin{lstlisting}[]
Please answer 1 if angioplasty or stent placement in the arterial system was performed. The arterial system is defined as any artery in the body. Answer 0 if angioplasty or stent placement in the arterial system was not performed or not documented
\end{lstlisting}

\texttt{\normalsize Prompt for Question  15:}
\begin{lstlisting}[]
Please answer 1 if angioplasty or stent placement in the venous system was performed. The venous system is defined as any vein in the body. Angioplasty means any inflation of a balloon in the venous system for opening a lumen or for mechanical disruption. Answer 0 if angioplasty or stent placement in the venous system was not performed or not documented
\end{lstlisting}

\texttt{\normalsize Prompt for Question  16:}
\begin{lstlisting}[]
Please answer 1 if stent placement in the carotid artery was performed and 0 if it was not performed or not documented
\end{lstlisting}

\texttt{\normalsize Prompt for Question  17:}
\begin{lstlisting}[]
Please answer 1 if thrombolytic therapy or thrombectomy was performed. Thrombolytic therapy may be administration of any tissue plasminogen activator (tPA) usch as tenecteplase or alteplase. Thrombectomy may include mechanical suction thrombectomy, balloon thrombectomy, or pharmacomechanical thrombectomy. Answer 0 if thrombolytic therapy or thrombectomy was not performed or not documented
\end{lstlisting}

\texttt{\normalsize Prompt for Question  18:}
\begin{lstlisting}[]
Please answer 1 if aortic endograft placement or revision was performed and 0 if it was not performed or not documented
\end{lstlisting}

\texttt{\normalsize Prompt for Question  19:}
\begin{lstlisting}[]
Please answer 1 if an emergency embolization was performed. Emergency embolization is defined as the intravascular administration of an occlusive material to stop or control bleeding and hemrrohage. Possible indications include trauma, gastrointestinal (GI) bleeding, hemoptysis, iatrogenic, and tumoral. Embolization materials may include Gelfoam, Onyx, beads, coils, and plugs. Answer 0 if an emergency embolization was not performed or not documented
\end{lstlisting}

\texttt{\normalsize Prompt for Question  20:}
\begin{lstlisting}[]
Please answer 1 if an elective embolization was performed. Elective embolization is defined as the intravascular administration of an occlusive material to stop or reduce flow of blood or for administration of a medication. Elective embolization procedures include uterine artery emboliztion (UAE), uterine fibroid embolization (UFE), arteriovenous malformation (AVM) embolization, varicocele embolization, and gondal vein embolization. Transarterial chemoembolization (TACE) and transarterial radioemboliztion (TARE) should not be included in this group unless protective embolization was performed. Answer 0 if an elective embolization was not performed or not documented
\end{lstlisting}

\texttt{\normalsize Prompt for Question  21:}
\begin{lstlisting}[]
Please answer 1 if a transarterial chemoembolization (TACE) was performed. TACE is defined as the intravascular administration of a chemotherapy gent to treat a tumor. Answer 0 if a TACE procedure was not performed or not documented
\end{lstlisting}

\texttt{\normalsize Prompt for Question  22:}
\begin{lstlisting}[]
Please answer 1 if a transarterial radioembolization (TARE) procedure was performed. TARE procedures include angiographic mapping with or without MAA administration and yttrium-90 (Y90) administration. Answer 0 if a TARE procedure was not performed or not documented
\end{lstlisting}

\texttt{\normalsize Prompt for Question  23:}
\begin{lstlisting}[]
Please answer 1 if an intravascular procedure was performed that doesn't fit into one of the following categories: Venous access (such as port placement, exchange, or removal, or central venous catheter placement, exchange, or removal, or dialysis catheter placement, exchange, or removal), dialysis access intervention, IVC filter placement or removal, ablation, stent or stent-graft placement, TIPS, DIPS, embolization, TACE, or TARE. Examples might include a transjugular liver or renal biopsy, other intravscalar biopsy or fiducial placement. Otherwise, please answer 0
\end{lstlisting}

\texttt{\normalsize Prompt for Question  24:}
\begin{lstlisting}[]
Please answer 1 if a percutaneous biopsy was performed and 0 if it was not performed or not documented. Percutaneous biopsy does not include intravascular biopsy procedures such as transjugular liver biopsy and transsjugular renal biopsy
\end{lstlisting}

\texttt{\normalsize Prompt for Question  25:}
\begin{lstlisting}[]
Please answer 1 if an abscess drainage procedure was performed. These procedures include an abscess or fluid collection drain placement, revision, repositioning, upsize, or removal. This excludes biliary drains and tubes, cholecystostomy drains and tubes, nephrostomy drains and tubes, nephroureterostomy drains and tubes, and chest drains and tubes. Answer 0 if an abscess drainage procedure was not performed or not documented
\end{lstlisting}

\texttt{\normalsize Prompt for Question  26:}
\begin{lstlisting}[]
Please answer 1 if  paracentesis or thoracentesis was performed and 0 if it was not performed or not documented
\end{lstlisting}

\texttt{\normalsize Prompt for Question  27:}
\begin{lstlisting}[]
Please answer 1 if a chest tube placement was performed and 0 if it was not performed or not documented
\end{lstlisting}

\texttt{\normalsize Prompt for Question  28:}
\begin{lstlisting}[]
Please answer 1 if pleurodesis was performed and 0 if it was not performed or not documented
\end{lstlisting}

\texttt{\normalsize Prompt for Question  29:}
\begin{lstlisting}[]
Please answer 1 if any procedure involving a biliary drain or stent was performed. These procedures include percutaneous transhepatic cholangiography (PTC), biliary drain placement, evaluation, or revision, and biliary stent placement or revision. Cholecystostomy placement/exchange/removal procedures are not considered a biliary drain or stent procedure. Answer 0 if no biliary drain or biliary stent procedure was performed or documented
\end{lstlisting}

\texttt{\normalsize Prompt for Question  30:}
\begin{lstlisting}[]
Please answer 1 if any procedure involving a geniturinary drain procedure was performed. These procedures include nephrostomy tube and nephrourteral stent placement, exchange, revision, or removal. Answer 0 if any procedure involving a geniturinary drain procedure was not performed or not documented
\end{lstlisting}

\texttt{\normalsize Prompt for Question  31:}
\begin{lstlisting}[]
Please answer 1 if any enterostomy procedure was performed. These procedures include gastrostomy, gastrojejunostomy, jejunostromy, iliostomy, and cecostomy placement, exchange, revision, or removal. Answer 0 if an enterostomy procedure was not performed or not documented
\end{lstlisting}

\texttt{\normalsize Prompt for Question  32:}
\begin{lstlisting}[]
Please answer 1 if any cholecystostomy procedure was performed. These procedures include cholecystostomy drain or tube placement, exchange, revision, or removal and cholecystography. Answer 0 if a cholecystostomy procedure was not performed or not documented
\end{lstlisting}

\texttt{\normalsize Prompt for Question  33:}
\begin{lstlisting}[]
Please answer 1 if a cyst or lymphocele procedure was performed. These procedures include aspiration, drain placement, exchange, revision, or removal and sclerosis of a cyst or lymphocele. This does not include interventions on abscesses or other fluid collections. Answer 0 if a cyst or lymphocele procedure was not performed or not documented
\end{lstlisting}

\texttt{\normalsize Prompt for Question  34:}
\begin{lstlisting}[]
Please answer 1 if a nonvascular stent placement was performed. Nonvascular stents include esophageal, tracheobronchial, duodenal, and colonic stents. Answer 0 if a nonvascular stent placement was not performed or not documented
\end{lstlisting}

\texttt{\normalsize Prompt for Question  35:}
\begin{lstlisting}[]
Please answer 1 if a miscellaneous nonvascular transplant intervention was performed. Miscellaneous nonvascular transplant interventions are interventions performed exclusively on a transplant organ outside of the arteries and veins. The text must explicitly report that the organ is a transplant organ. A transplant organ is not a native organ. Intravascular procedures involving a transplant organ, such as transplant renal artery stenosis (TRAS) or transjular biopsy, should not considered as examples because they are vascular. Examples of nonvascular transplant interventions include stent placements involving a transplant organ or balloon plasty procedures such as ureteroplasty. Procedures such as nephrostomy and nephroureterostomy placement, exchange, or removal and biliary tube placement, exchange, and removal should NOT be considered in this miscellaneous category. Answer 0 if a miscellaneous nonvascular transplant intervention was not performed or not documented
\end{lstlisting}

\texttt{\normalsize Prompt for Question  36:}
\begin{lstlisting}[]
Please answer 1 if a tumor ablation was performed. Tumor ablations may include radiofrequency (RFA), laser, microwave, cryoablation, and ethanol administration. Answer 0 if a tumor ablation was not performed or not documented
\end{lstlisting}

\texttt{\normalsize Prompt for Question  37:}
\begin{lstlisting}[]
Please answer 1 if a pain managment procedure was performed. Pain managment procedures may include steroid injection, celiac plexus neurolysis, or nerve and periosteal cryoablation. Answer 0 if a pain managment procedure was not performed or not documented
\end{lstlisting}

\texttt{\normalsize Prompt for Question  38:}
\begin{lstlisting}[]
Please answer 1 if a Fallopian tube recanalization was performed and 0 if it was not performed or not documented
\end{lstlisting}

\texttt{\normalsize Prompt for Question  39:}
\begin{lstlisting}[]
An invasive procedure involves inserting instruments or devices into the body, typically through the skin or a natural body orifice. Please answer 1 if the report specifies that a nonvascular, invasive procedure was performed and it does not fall into any of the following categories: Biopsy, Abscess drainage & tube checks, Paracentesis, Thoracentesis, Chest tube placement, Pleurodesis, PTC, Biliary drainage catheter placement/exchange/removal, Biliary stent placement/exchange/removal, Tube checks, Genitourinary catheter placement/exchange/removal, Nephrostomy placement/exchange/removal, Nephroureteral tube placement/exchange/removal, Gastrostomy placement/exchange/removal, Gastrojejunostomy placement/exchange/removal, Cholecystostomy placement/exchange/removal, Cyst or lymphocele intervention, Nonvascular stents (esophageal, tracheobronchial, duodenal, colonic), Miscellaneous transplant interventions, Tumor ablation, Pain management interventions, Fallopian tube recanalization, or any intravascular procedure (including venous access procedures such as port placement, exchange, or removal; central venous catheter placement, exchange, or removal; dialysis catheter placement, exchange, or removal; dialysis access intervention; IVC filter placement or removal; Ablation; Stent or stent-graft placement; TIPS; DIPS; Embolization; Transarterial chemoembolization (TACE); or Transarterial radioembolization (TARE)). If a non-vascuar invasive procedure was not performed, please answer 0.
\end{lstlisting}

\subsection{Chain-of-Thought}

\texttt{\normalsize Prompt for Question  1:}
\begin{lstlisting}[]
Your task is to determine whether a **CTA (CT Angiography)** or **intra-operative cone-beam CT with contrast** was explicitly performed.  

If the procedure was explicitly performed, return `1` along with an explanation. Otherwise, return `0` with an explanation.

Follow a structured reasoning approach to do the task:

### **Reasoning Steps**  
1. **Identify Mentions**: Extract all references to CTA or intra-operative cone-beam CT with contrast in the report.  
2. **Assess Explicitness**: Determine if the report explicitly states that the procedure **was performed** (e.g., `CTA was conducted,' 'A cone-beam CT with contrast was completed').  
3. **Resolve Ambiguities**: If the procedure is only **suggested, planned, considered, or recommended**, but there is no explicit confirmation that it was performed, classify it as **not performed**.  
4. **Handle Uncertainty**: If the report lacks a direct mention of the procedure, assume it was **not performed**.  
5. **Generate JSON Output**: Construct a response in strict JSON format, adhering to the following structure:  
\end{lstlisting}

\texttt{\normalsize Prompt for Question  2:}
\begin{lstlisting}[]

Your task is to determine whether an **MRA (MR Angiography)** was explicitly performed. 

If the procedure was explicitly performed, return `1` along with an explanation. Otherwise, return `0` with an explanation.

Follow a structured reasoning approach to do the task:

### **Reasoning Steps**  
1. **Identify Mentions**: Extract all references to MRA (MR Angiography) in the report.  
2. **Assess Explicitness**: Determine if the report explicitly states that the procedure **was performed** (e.g., 'MRA was conducted,' 'MR Angiography was completed').  
3. **Resolve Ambiguities**: If the procedure is only **suggested, planned, considered, or recommended**, but there is no explicit confirmation that it was performed, classify it as **not performed**.  
4. **Handle Uncertainty**: If the report lacks a direct mention of MRA, assume it was **not performed**.  
5. **Generate JSON Output**: Construct a response in strict JSON format, adhering to the following structure:  
\end{lstlisting}

\texttt{\normalsize Prompt for Question  3:}
\begin{lstlisting}[]

Your task is to determine whether a **noninvasive vascular lab study** was explicitly performed. Examples include **duplex ultrasound, color flow study, or Pulse Volume Recording (PVR)**. 

If the procedure was explicitly performed, return `1` along with an explanation. Otherwise, return `0` with an explanation.

Follow a structured reasoning approach to do the task:

### **Reasoning Steps**  
1. **Identify Mentions**: Extract all references to **duplex ultrasound, color flow study, Pulse Volume Recording (PVR), or any other noninvasive vascular lab study** in the report.  
2. **Assess Explicitness**: Determine if the report clearly states that the procedure **was performed** (e.g., 'A duplex ultrasound was conducted,' 'PVR testing was completed').  
3. **Resolve Ambiguities**: If the procedure is only **suggested, planned, considered, or recommended** but not explicitly confirmed as performed, classify it as **not performed**.  
4. **Handle Uncertainty**: If the report lacks a direct mention of a noninvasive vascular lab study, assume it **was not performed**.  
5. **Generate JSON Output**: Construct a response in strict JSON format, adhering to the following structure:  

\end{lstlisting}

\texttt{\normalsize Prompt for Question  4:}
\begin{lstlisting}[]

Your task is to determine whether a **cardiac imaging study** was explicitly performed. Examples include a **CT coronary angiogram** or **nuclear medicine cardiac stress test**. 

If the procedure was explicitly performed, return `1` along with an explanation. Otherwise, return `0` with an explanation.

Follow a structured reasoning approach to do the task:

### **Reasoning Steps**  
1. **Identify Mentions**: Extract all references to **CT coronary angiogram, nuclear medicine cardiac stress test, or any other cardiac imaging study** in the report.  
2. **Assess Explicitness**: Determine if the report clearly states that the procedure **was performed** (e.g., 'A CT coronary angiogram was conducted,' 'A nuclear medicine cardiac stress test was completed').  
3. **Resolve Ambiguities**: If the procedure is only **suggested, planned, considered, or recommended** but not explicitly confirmed as performed, classify it as **not performed**.  
4. **Handle Uncertainty**: If the report lacks a direct mention of a cardiac imaging study, assume it **was not performed**.  
5. **Generate JSON Output**: Construct a response in strict JSON format, adhering to the following structure:  
\end{lstlisting}

\texttt{\normalsize Prompt for Question  5:}
\begin{lstlisting}[]

Your task is to determine whether a **dedicated arteriography** was explicitly performed. **Arteriography** is defined as imaging of arteries using **fluoroscopy and contrast** specifically to evaluate arterial anatomy or pathology. The contrast **must be administered directly into an artery**. **Venography does not count as arteriography.** 

If the procedure was explicitly performed, return `1` along with an explanation. Otherwise, return `0` with an explanation.

Follow a structured reasoning approach to do the task:

### **Reasoning Steps**  
1. **Identify Mentions**: Extract all references to **arteriography** in the report. Look for terms like **arteriogram, fluoroscopic arterial imaging, or direct intra-arterial contrast administration**.  
2. **Confirm Modality & Contrast Administration**: Ensure the report specifies that the imaging was performed **using fluoroscopy and contrast**. The contrast **must** be administered **into an artery** to qualify as arteriography.  
3. **Differentiate from Venography**: If the procedure involves **venous imaging** (e.g., venography or venogram), **do not classify it as arteriography**.  
4. **Resolve Ambiguities**: If the procedure is only **suggested, planned, considered, or recommended** but not explicitly confirmed as performed, classify it as **not performed**.  
5. **Handle Uncertainty**: If the report lacks a direct mention of **arteriography with intra-arterial contrast administration**, assume it **was not performed**.  
6. **Generate JSON Output**: Construct a response in strict JSON format, adhering to the following structure:  
\end{lstlisting}

\texttt{\normalsize Prompt for Question  6:}
\begin{lstlisting}[]

Your task is to determine whether a **dedicated venography** was explicitly performed. **Venography** is defined as imaging of veins using **fluoroscopy and contrast** specifically to evaluate **venous anatomy or pathology**. The contrast **must be administered directly into a vein**. **Arteriography does not count as venography.** 

If the procedure was explicitly performed, return `1` along with an explanation. Otherwise, return `0` with an explanation.

Follow a structured reasoning approach to do the task:

### **Reasoning Steps**  
1. **Identify Mentions**: Extract all references to **venography** in the report. Look for terms like **venogram, fluoroscopic venous imaging, or direct intra-venous contrast administration**.  
2. **Confirm Modality & Contrast Administration**: Ensure the report specifies that the imaging was performed **using fluoroscopy and contrast**. The contrast **must** be administered **into a vein** to qualify as venography.  
3. **Differentiate from Arteriography**: If the procedure involves **arterial imaging** (e.g., arteriography or arteriogram), **do not classify it as venography**.  
4. **Resolve Ambiguities**: If the procedure is only **suggested, planned, considered, or recommended** but not explicitly confirmed as performed, classify it as **not performed**.  
5. **Handle Uncertainty**: If the report lacks a direct mention of a **venography with intra-venous contrast administration**, assume it **was not performed**.  
6. **Generate JSON Output**: Construct a response in strict JSON format, adhering to the following structure:  
\end{lstlisting}

\texttt{\normalsize Prompt for Question  7:}
\begin{lstlisting}[]

Your task is to determine whether a **dialysis access evaluation** was explicitly performed. These evaluations include:  
- **Ultrasound studies** (e.g., Doppler ultrasound of dialysis access).  
- **Fistulagrams** (fluoroscopic imaging of a dialysis fistula using contrast).  
- **Graft evaluations** (imaging studies assessing the function and patency of dialysis grafts).  

If the procedure was explicitly performed, return `1` along with an explanation. Otherwise, return `0` with an explanation.

Follow a structured reasoning approach to do the task:

### **Reasoning Steps**  
1. **Identify Mentions**: Extract all references to **dialysis access evaluation** in the report. Look for terms such as **dialysis access ultrasound, fistulagram, or graft evaluation**.  
2. **Confirm the Type of Study**: Ensure that the report explicitly states that one of the following procedures was performed:  
   - **Ultrasound study** for dialysis access.  
   - **Fistulagram** (contrast-based evaluation of a fistula).  
   - **Graft evaluation** (assessment of a dialysis graft using imaging).  
3. **Resolve Ambiguities**: If the procedure is only **suggested, planned, considered, or recommended** but not explicitly confirmed as performed, classify it as **not performed**.  
4. **Handle Uncertainty**: If the report lacks a direct mention of any of the qualifying dialysis access studies, assume it **was not performed**.  
5. **Generate JSON Output**: Construct a response in strict JSON format, adhering to the following structure:  

\end{lstlisting}

\texttt{\normalsize Prompt for Question  8:}
\begin{lstlisting}[]

Your task is to determine whether **carotid artery imaging** was explicitly performed. This includes imaging studies specifically evaluating the carotid arteries, such as:  
- **Carotid ultrasound (Doppler or duplex ultrasound of the carotid arteries)**  
- **Carotid CT angiography (CTA)**  
- **Carotid MR angiography (MRA)**  
- **Carotid angiography (fluoroscopy with contrast administered into the carotid arteries)**  

If the procedure was explicitly performed, return `1` along with an explanation. Otherwise, return `0` with an explanation.

Follow a structured reasoning approach to do the task:

### **Reasoning Steps**  
1. **Identify Mentions**: Extract all references to **carotid artery imaging** in the report. Look for terms such as **carotid ultrasound, carotid Doppler, carotid CTA, carotid MRA, or carotid angiogram**.  
2. **Confirm the Imaging Modality**: Ensure that the report explicitly states that one of the following procedures was performed:  
   - **Ultrasound study** (e.g., Doppler or duplex of the carotid arteries).  
   - **CT angiography (CTA)** of the carotid arteries.  
   - **MR angiography (MRA)** of the carotid arteries.  
   - **Catheter-based carotid angiography** (contrast-enhanced fluoroscopic evaluation).  
3. **Resolve Ambiguities**: If the procedure is only **suggested, planned, considered, or recommended** but not explicitly confirmed as performed, classify it as **not performed**.  
4. **Handle Uncertainty**: If the report lacks a direct mention of any carotid artery imaging study, assume it **was not performed**.  
5. **Generate JSON Output**: Construct a response in strict JSON format, adhering to the following structure:  

\end{lstlisting}

\texttt{\normalsize Prompt for Question  9:}
\begin{lstlisting}[]

Your task is to determine whether a **central venous catheter (CVC) procedure** was explicitly performed. This includes:  
- **Placement** of a central venous catheter, port, or peripherally inserted central catheter (PICC).  
- **Removal** of a central venous catheter, port, or PICC.  
- **Revision** or adjustment of an existing central venous catheter, port, or PICC.  

If the procedure was explicitly performed, return `1` along with an explanation. Otherwise, return `0` with an explanation.

Follow a structured reasoning approach to do the task:

### **Reasoning Steps**  
1. **Identify Mentions**: Extract all references to **central venous catheter procedures** in the report. Look for terms such as **CVC placement, central line insertion, port-a-cath placement, PICC line insertion, catheter revision, or catheter removal**.  
2. **Confirm Procedure Type**: Ensure that the report explicitly states that one of the following procedures was performed:  
   - **Placement of a CVC, port, or PICC** (e.g., via ultrasound or fluoroscopic guidance).  
   - **Removal of a CVC, port, or PICC** (e.g., catheter extraction or explantation).  
   - **Revision or adjustment** of an existing CVC, port, or PICC.  
3. **Resolve Ambiguities**: If the procedure is only **suggested, planned, considered, or recommended** but not explicitly confirmed as performed, classify it as **not performed**.  
4. **Handle Uncertainty**: If the report lacks a direct mention of any **CVC-related procedure**, assume it **was not performed**.  
5. **Generate JSON Output**: Construct a response in strict JSON format, adhering to the following structure:  
\end{lstlisting}

\texttt{\normalsize Prompt for Question  10:}
\begin{lstlisting}[]

Your task is to determine whether an **Inferior Vena Cava (IVC) filter placement or removal procedure** was explicitly performed. This includes:  
- **Placement** of an IVC filter (implantation of a filter within the inferior vena cava to prevent embolism).  
- **Removal** or **retrieval** of an IVC filter (extraction of a previously placed filter).  

If the procedure was explicitly performed, return `1` along with an explanation. Otherwise, return `0` with an explanation.

Follow a structured reasoning approach to do the task:

### **Reasoning Steps**  
1. **Identify Mentions**: Extract all references to **IVC filter procedures** in the report. Look for terms such as **IVC filter placement, IVC filter removal, IVC filter retrieval, or caval filter implantation**.  
2. **Confirm Procedure Type**: Ensure that the report explicitly states that one of the following procedures was performed:  
   - **Placement of an IVC filter** (e.g., via catheter-based insertion in the inferior vena cava).  
   - **Removal or retrieval of an existing IVC filter** (e.g., endovascular retrieval of the device).  
3. **Resolve Ambiguities**: If the procedure is only **suggested, planned, considered, or recommended** but not explicitly confirmed as performed, classify it as **not performed**.  
4. **Handle Uncertainty**: If the report lacks a direct mention of any **IVC filter placement or removal**, assume it **was not performed**.  
5. **Generate JSON Output**: Construct a response in strict JSON format, adhering to the following structure:  

\end{lstlisting}

\texttt{\normalsize Prompt for Question  11:}
\begin{lstlisting}[]

Your task is to determine whether a **venous ablation procedure** was explicitly performed. Venous ablations are performed primarily to treat **varicose veins** and involve techniques such as:  
- **Endovenous laser ablation (EVLA)**  
- **Radiofrequency ablation (RFA)**  
- **Chemical sclerotherapy (foam or liquid injection for vein closure)**  
- **Mechanochemical ablation (MOCA, ClariVein, etc.)**  

If the procedure was explicitly performed, return `1` along with an explanation. Otherwise, return `0` with an explanation.

Follow a structured reasoning approach to do the task:

### **Reasoning Steps**  
1. **Identify Mentions**: Extract all references to **venous ablation procedures** in the report. Look for terms such as **venous ablation, endovenous laser ablation (EVLA), radiofrequency ablation (RFA), sclerotherapy, or mechanochemical ablation**.  
2. **Confirm the Procedure Type**: Ensure that the report explicitly states that one of the following procedures was performed:  
   - **Endovenous thermal ablation (laser or radiofrequency-based techniques).**  
   - **Sclerotherapy (chemical injection for vein closure).**  
   - **Mechanochemical ablation (combining mechanical and chemical vein closure techniques).**  
3. **Resolve Ambiguities**: If the procedure is only **suggested, planned, considered, or recommended** but not explicitly confirmed as performed, classify it as **not performed**.  
4. **Handle Uncertainty**: If the report lacks a direct mention of any **venous ablation procedure**, assume it **was not performed**.  
5. **Generate JSON Output**: Construct a response in strict JSON format, adhering to the following structure:  

\end{lstlisting}

\texttt{\normalsize Prompt for Question  12:}
\begin{lstlisting}[]

Your task is to determine whether a **dialysis access intervention** was explicitly performed. These interventions are specific to **arteriovenous (AV) access** and include:  
- **Fistulagrams** (angiographic evaluation of an AV fistula).  
- **Graft evaluations** (imaging or interventional assessment of AV grafts).  
- **Declot procedures** (thrombectomy or thrombolysis specifically performed on AV access).  

If the procedure was explicitly performed, return `1` along with an explanation. Otherwise, return `0` with an explanation.

Follow a structured reasoning approach to do the task:

### **Reasoning Steps**  
1. **Identify Mentions**: Extract all references to **dialysis access interventions** in the report. Look for terms such as **fistulagram, AV fistula angiography, AV graft evaluation, thrombectomy, thrombolysis, or declot procedure on an AV access**.  
2. **Confirm the Procedure Type**: Ensure that the report explicitly states that one of the following procedures was performed **on an arteriovenous access**:  
   - **Fistulagram** (AV fistula imaging and assessment).  
   - **Graft evaluation** (assessment of an AV graft for function or complications).  
   - **Declot procedure** (thrombectomy or thrombolysis **specifically on an AV access**).  
3. **Resolve Ambiguities**: If the procedure is only **suggested, planned, considered, or recommended** but not explicitly confirmed as performed, classify it as **not performed**.  
4. **Exclude Non-Relevant Procedures**: Ensure that procedures **not related to arteriovenous access** (such as central venous catheter interventions) **are not mistakenly classified as dialysis access interventions**.  
5. **Handle Uncertainty**: If the report lacks a direct mention any **dialysis access intervention**, assume it **was not performed**.  
6. **Generate JSON Output**: Construct a response in strict JSON format, adhering to the following structure:  

\end{lstlisting}

\texttt{\normalsize Prompt for Question  13:}
\begin{lstlisting}[]

Your task is to determine whether **any intervention involving a portosystemic shunt** was explicitly performed. These interventions include:  
- **Shunt placement** (creation of a new portosystemic shunt).  
- **Shunt evaluation** (imaging or assessment of an existing shunt for patency or function).  
- **Shunt revision** (modification, balloon angioplasty, or stenting of a pre-existing shunt).  

The **portosystemic shunts** of interest include:  
- **Transjugular Intrahepatic Portosystemic Shunt (TIPS)**  
- **Direct Intrahepatic Portocaval Shunt (DIPS)**  
- **Transjugular Transcaval Intrahepatic Portosystemic Shunt (TTIPS)**  

If the procedure was explicitly performed, return `1` along with an explanation. Otherwise, return `0` with an explanation.

Follow a structured reasoning approach to do the task:

### **Reasoning Steps**  
1. **Identify Mentions**: Extract all references to **portosystemic shunt interventions** in the report. Look for terms such as **TIPS, DIPS, TTIPS, transjugular intrahepatic shunt, portocaval shunt, portosystemic shunt placement, shunt revision, or shunt evaluation**.  
2. **Confirm the Procedure Type**: Ensure that the report explicitly states that one of the following procedures was performed:  
   - **Shunt Placement** (creation of a new TIPS, DIPS, or TTIPS).  
   - **Shunt Evaluation** (assessment of an existing shunt's function, flow, or patency).  
   - **Shunt Revision** (angioplasty, stenting, or any modification to an existing shunt).  
3. **Resolve Ambiguities**: If the procedure is only **suggested, planned, considered, or recommended** but not explicitly confirmed as performed, classify it as **not performed**.  
4. **Exclude Non-Relevant Procedures**: Ensure that unrelated vascular or hepatic interventions (such as portal vein thrombectomy or general liver angiography) **are not mistaken for portosystemic shunt interventions**.  
5. **Handle Uncertainty**: If the report lacks a direct mention of any **portosystemic shunt intervention**, assume it **was not performed**.  
6. **Generate JSON Output**: Construct a response in strict JSON format, adhering to the following structure:  

\end{lstlisting}

\texttt{\normalsize Prompt for Question  14:}
\begin{lstlisting}[]

Your task is to determine whether **angioplasty or stent placement in the arterial system** was explicitly performed. The **arterial system** includes **any artery in the body**, such as:  
- **Coronary arteries** (e.g., coronary angioplasty, coronary stenting).  
- **Carotid arteries** (e.g., carotid artery stenting, carotid angioplasty).  
- **Peripheral arteries** (e.g., femoral, iliac, popliteal artery interventions).  
- **Visceral arteries** (e.g., renal, mesenteric artery stenting).  
- **Aortic branches** (e.g., subclavian, vertebral artery stenting).  

If the procedure was explicitly performed, return `1` along with an explanation. Otherwise, return `0` with an explanation.

Follow a structured reasoning approach to do the task:

### **Reasoning Steps**  
1. **Identify Mentions**: Extract all references to **angioplasty or stent placement** within the **arterial system** from the report. Look for terms such as **angioplasty, percutaneous transluminal angioplasty (PTA), arterial stenting, vascular stenting, endovascular stent placement, or balloon dilation of an artery**.  
2. **Confirm the Procedure Type**: Ensure that the report explicitly states that one of the following procedures was performed:  
   - **Angioplasty** (balloon dilation of an arterial stenosis).  
   - **Stent Placement** (implantation of a stent within an artery).  
3. **Resolve Ambiguities**: If the procedure is only **suggested, planned, considered, or recommended** but not explicitly confirmed as performed, classify it as **not performed**.  
4. **Exclude Non-Relevant Procedures**: Ensure that interventions performed **in the venous system** (such as venous angioplasty or venous stenting) **are not mistaken for arterial interventions**.  
5. **Handle Uncertainty**: If the report lacks a direct mention of any **arterial angioplasty or stent placement**, assume it **was not performed**.  
6. **Generate JSON Output**: Construct a response in strict JSON format, adhering to the following structure:  

\end{lstlisting}

\texttt{\normalsize Prompt for Question  15:}
\begin{lstlisting}[]

Your task is to determine whether **angioplasty or stent placement in the venous system** was explicitly performed. The **venous system** includes **any vein in the body**, such as:  
- **Central veins** (e.g., superior vena cava (SVC), inferior vena cava (IVC), subclavian, brachiocephalic veins).  
- **Peripheral veins** (e.g., iliac, femoral, popliteal, upper extremity veins).  
- **Portal or hepatic veins** (e.g., hepatic vein stenting, portal vein angioplasty).  
- **Pelvic veins** (e.g., gonadal, ovarian, or internal iliac vein interventions).  

If the procedure was explicitly performed, return `1` along with an explanation. Otherwise, return `0` with an explanation.

Follow a structured reasoning approach to do the task:

### **Reasoning Steps**  
1. **Identify Mentions**: Extract all references to **venous angioplasty or venous stent placement** from the report. Look for terms such as **venous angioplasty, balloon venoplasty, venous stenting, endovenous stent placement, mechanical luminal disruption, or vein dilation with a balloon**.  
2. **Confirm the Procedure Type**: Ensure that the report explicitly states that one of the following procedures was performed:  
   - **Venous Angioplasty** (balloon inflation to open a vein or disrupt an obstruction).  
   - **Venous Stent Placement** (implantation of a stent within a vein).  
3. **Resolve Ambiguities**: If the procedure is only **suggested, planned, considered, or recommended** but not explicitly confirmed as performed, classify it as **not performed**.  
4. **Exclude Non-Relevant Procedures**: Ensure that interventions performed **in the arterial system** (such as arterial angioplasty or arterial stenting) **are not mistakenly classified as venous interventions**.  
5. **Handle Uncertainty**: If the report lacks direct mention of any **venous angioplasty or stent placement**, assume it **was not performed**.  
6. **Generate JSON Output**: Construct a response in strict JSON format, adhering to the following structure:  

\end{lstlisting}

\texttt{\normalsize Prompt for Question  16:}
\begin{lstlisting}[]

Your task is to determine whether **stent placement in the carotid artery** was explicitly performed. The carotid arteries include:  
- **Common carotid artery (CCA)**  
- **Internal carotid artery (ICA)**  
- **External carotid artery (ECA)**  

If the procedure was explicitly performed, return `1` along with an explanation. Otherwise, return `0` with an explanation.

Follow a structured reasoning approach to do the task:

### **Reasoning Steps**  
1. **Identify Mentions**: Extract all references to **stent placement in the carotid artery** from the report. Look for terms such as **carotid stenting, carotid artery stent placement, endovascular stenting of the carotid, carotid angioplasty with stenting, or CAS (Carotid Artery Stenting)**.  
2. **Confirm the Procedure Type**: Ensure that the report explicitly states that **a stent was placed in the carotid artery**.  
3. **Resolve Ambiguities**: If the procedure is only **suggested, planned, considered, or recommended** but not explicitly confirmed as performed, classify it as **not performed**.  
4. **Exclude Non-Relevant Procedures**: Ensure that other **carotid interventions (such as carotid angiography, balloon angioplasty without stent placement, or diagnostic imaging)** are **not mistaken for carotid stenting**.  
5. **Handle Uncertainty**: If the report lacks a direct mention of **carotid artery stent placement**, assume it **was not performed**.  
6. **Generate JSON Output**: Construct a response in strict JSON format, adhering to the following structure:  

\end{lstlisting}

\texttt{\normalsize Prompt for Question  17:}
\begin{lstlisting}[]

Your task is to determine whether **thrombolytic therapy or thrombectomy** was explicitly performed. These interventions include:  

#### **Thrombolytic Therapy** (Clot-Dissolving Medication)  
- **Systemic or catheter-directed administration** of thrombolytic agents.  
- **Common drugs:**  
  - **tPA (tissue plasminogen activator)**  
  - **Tenecteplase**  
  - **Alteplase**  

#### **Thrombectomy** (Clot Removal Procedures)  
- **Mechanical thrombectomy** - Direct clot removal using specialized devices (e.g., aspiration catheters, stent retrievers).  
- **Balloon thrombectomy** - Clot extraction using a balloon catheter.  
- **Pharmacomechanical thrombectomy** - Combination of thrombolytic drugs with mechanical clot disruption.  

If the procedure was explicitly performed, return `1` along with an explanation. Otherwise, return `0` with an explanation.

Follow a structured reasoning approach to do the task:

### **Reasoning Steps**  
1. **Identify Mentions**: Extract all references to **thrombolytic therapy or thrombectomy** from the report. Look for terms such as **tPA administration, thrombolysis, lytic therapy, alteplase, tenecteplase, mechanical thrombectomy, aspiration thrombectomy, balloon thrombectomy, or pharmacomechanical thrombectomy**.  
2. **Confirm the Procedure Type**: Ensure that the report explicitly states that **one of the following was performed**:  
   - **Thrombolytic therapy** (administration of clot-dissolving drugs).  
   - **Thrombectomy** (mechanical or pharmacomechanical clot removal).  
3. **Resolve Ambiguities**: If the procedure is only **suggested, planned, considered, or recommended** but not explicitly confirmed as performed, classify it as **not performed**.  
4. **Exclude Non-Relevant Procedures**: Ensure that other **vascular interventions (such as angioplasty or stenting)** are **not mistaken for thrombolytic therapy or thrombectomy**.  
5. **Handle Uncertainty**: If the report lacks a direct mention of **thrombolytic therapy or thrombectomy**, assume it **was not performed**.  
6. **Generate JSON Output**: Construct a response in strict JSON format, adhering to the following structure:  

\end{lstlisting}

\texttt{\normalsize Prompt for Question  18:}
\begin{lstlisting}[]

Your task is to determine whether **aortic endograft placement or revision** was explicitly performed. This includes procedures related to:  

- **Endovascular Aneurysm Repair (EVAR)** - Placement of an endograft to treat an abdominal aortic aneurysm (AAA).  
- **Thoracic Endovascular Aortic Repair (TEVAR)** - Endograft placement in the thoracic aorta.  
- **Fenestrated Endovascular Aortic Repair (FEVAR)** - Endograft placement with fenestrations for branch vessels.  
- **Endograft Revision** - Any modification, extension, or repair of a previously placed aortic endograft.  

If the procedure was explicitly performed, return `1` along with an explanation. Otherwise, return `0` with an explanation.

Follow a structured reasoning approach to do the task:

### **Reasoning Steps**  
1. **Identify Mentions**: Extract all references to **aortic endograft placement or revision** from the report. Look for terms such as **EVAR, TEVAR, FEVAR, aortic stent graft placement, aortic endograft deployment, endovascular aortic repair, endograft revision, or endograft extension**.  
2. **Confirm the Procedure Type**: Ensure that the report explicitly states that **one of the following was performed**:  
   - **Aortic endograft placement** (initial deployment of an aortic stent graft).  
   - **Aortic endograft revision** (modification or extension of a prior graft).  
3. **Resolve Ambiguities**: If the procedure is only **suggested, planned, considered, or recommended** but not explicitly confirmed as performed, classify it as **not performed**.  
4. **Exclude Non-Relevant Procedures**: Ensure that other **vascular interventions (such as aortic angiography, open surgical repair, or stenting of non-aortic vessels)** are **not mistaken for aortic endograft placement or revision**.  
5. **Handle Uncertainty**: If the report lacks a direct mention of **aortic endograft placement or revision**, assume it **was not performed**.  
6. **Generate JSON Output**: Construct a response in strict JSON format, adhering to the following structure:  

\end{lstlisting}

\texttt{\normalsize Prompt for Question  19:}
\begin{lstlisting}[]

Your task is to determine whether **an emergency embolization** was explicitly performed. Emergency embolization is defined as the **intravascular administration of an occlusive material to stop or control active bleeding or hemorrhage**.  

#### **Common Indications for Emergency Embolization**  
- **Trauma-related bleeding** (e.g., liver, spleen, kidney, or pelvic trauma).  
- **Gastrointestinal (GI) bleeding** (e.g., gastric, duodenal, or colonic bleeding).  
- **Hemoptysis** (severe lung bleeding).  
- **Iatrogenic bleeding** (bleeding caused by medical procedures).  
- **Tumor-related hemorrhage** (e.g., bleeding from hypervascular tumors).  

#### **Common Embolization Materials**  
- **Gelfoam** (temporary occlusion).  
- **Onyx** (liquid embolic agent).  
- **Beads or microparticles** (used for tumor or GI bleeding control).  
- **Coils and plugs** (mechanical occlusion devices).  

If the procedure was explicitly performed, return `1` along with an explanation. Otherwise, return `0` with an explanation.

Follow a structured reasoning approach to do the task:

### **Reasoning Steps**  
1. **Identify Mentions**: Extract all references to **emergency embolization** in the report. Look for terms such as **embolization, endovascular embolization, vascular occlusion, coil embolization, Onyx injection, Gelfoam, particle embolization, or hemorrhage control procedure**.  
2. **Confirm the Procedure Type**: Ensure that the report explicitly states that an **embolization procedure** was performed **for emergency bleeding control** rather than elective or prophylactic purposes.  
3. **Determine the Indication**: Verify that the embolization was performed for **active bleeding, trauma, GI hemorrhage, hemoptysis, iatrogenic causes, or tumor-related bleeding** rather than for **tumor devascularization, arteriovenous malformation (AVM) treatment, or preoperative embolization**.  
4. **Resolve Ambiguities**: If the report mentions **embolization was considered, planned, or recommended** but does not confirm that it was **performed**, classify it as **not performed**.  
5. **Exclude Non-Relevant Procedures**: Ensure that **angiography, stent placement, thrombectomy, or other vascular interventions** are **not mistaken for emergency embolization**.  
6. **Handle Uncertainty**: If the report lacks a direct mention of **emergency embolization**, assume it **was not performed**.  
7. **Generate JSON Output**: Construct a response in strict JSON format, adhering to the following structure:  

\end{lstlisting}

\texttt{\normalsize Prompt for Question  20:}
\begin{lstlisting}[]

Your task is to determine whether **an elective embolization** was explicitly performed. Elective embolization is defined as the **intravascular administration of an occlusive material to stop or reduce blood flow for non-emergency therapeutic purposes** or for the **administration of medication**.  

#### **Types of Elective Embolization Procedures**  
- **Uterine Artery Embolization (UAE)** - Used to treat uterine fibroids.  
- **Uterine Fibroid Embolization (UFE)** - Specific type of UAE for fibroid treatment.  
- **Arteriovenous Malformation (AVM) Embolization** - Used to block abnormal blood vessel connections.  
- **Varicocele Embolization** - Treats enlarged veins in the scrotum.  
- **Gonadal Vein Embolization** - Treats venous insufficiency in reproductive organs.  
- **Protective Embolization** - Performed before procedures to prevent complications.  

#### **Exclusion Criteria**  
- **Transarterial Chemoembolization (TACE) and Transarterial Radioembolization (TARE)** should **not** be considered **elective embolization** **unless protective embolization** was performed.  
- **Emergency embolization for active bleeding or trauma** is **not elective** and should be excluded.  

If the procedure was explicitly performed, return `1` along with an explanation. Otherwise, return `0` with an explanation.

Follow a structured reasoning approach to do the task:

### **Reasoning Steps**  
1. **Identify Mentions**: Extract all references to **elective embolization** from the report. Look for terms such as **uterine artery embolization, fibroid embolization, AVM embolization, varicocele embolization, gonadal vein embolization, protective embolization, or planned vascular occlusion**.  
2. **Confirm the Procedure Type**: Ensure that the embolization was performed **for elective (non-emergency) indications**, such as **fibroids, AVMs, or varicoceles**, rather than for **acute hemorrhage control**.  
3. **Differentiate from Non-Elective Procedures**:  
   - **Exclude emergency embolization** for trauma, GI bleeding, hemoptysis, or tumor-related hemorrhage.  
   - **Exclude TACE or TARE unless protective embolization was explicitly mentioned.**  
4. **Resolve Ambiguities**: If the report mentions **embolization was considered, planned, or recommended** but does not confirm that it was **performed**, classify it as **not performed**.  
5. **Exclude Non-Relevant Procedures**: Ensure that **angiography, venous sclerotherapy, or other vascular interventions** are **not mistaken for elective embolization**.  
6. **Handle Uncertainty**: If the report lacks a direct mention of **elective embolization**, assume it **was not performed**.  
7. **Generate JSON Output**: Construct a response in strict JSON format, adhering to the following structure:  

\end{lstlisting}

\texttt{\normalsize Prompt for Question  21:}
\begin{lstlisting}[]

Your task is to determine whether **a transarterial chemoembolization (TACE) procedure** was explicitly performed. **TACE is defined as the intravascular administration of a chemotherapy agent to treat a tumor.**  

#### **Key Criteria for TACE**  
- **Must involve intra-arterial chemotherapy delivery** (e.g., doxorubicin, cisplatin, mitomycin C).  
- **Must involve arterial embolization** to reduce tumor blood supply.  
- **Indications typically include hepatocellular carcinoma (HCC) or metastatic liver tumors.**  

#### **Exclusion Criteria**  
- **Transarterial Radioembolization (TARE)** - Uses radioactive microspheres instead of chemotherapy.  
- **Elective Embolization (e.g., uterine fibroid embolization, AVM embolization)** - Not the same as TACE.  
- **Emergency Embolization** - Performed for bleeding, not tumor treatment.  

If the procedure was explicitly performed, return `1` along with an explanation. Otherwise, return `0` with an explanation.

Follow a structured reasoning approach to do the task:
### **Reasoning Steps**  
1. **Identify Mentions**: Extract all references to **TACE, transarterial chemoembolization, intra-arterial chemotherapy, or embolization for tumor treatment** in the report.  
2. **Confirm the Procedure Type**: Ensure the report explicitly states that **chemotherapy was administered intra-arterially** and that **embolization was performed as part of the procedure**.  
3. **Differentiate from Other Procedures**:  
   - **Exclude TARE** (radioembolization with Y-90 microspheres).  
   - **Exclude elective embolization for fibroids, AVMs, or varicoceles.**  
   - **Exclude emergency embolization for trauma or hemorrhage.**  
4. **Resolve Ambiguities**: If the report mentions **TACE was considered, planned, or recommended** but does not confirm that it was **performed**, classify it as **not performed**.  
5. **Exclude Non-Relevant Procedures**: Ensure that **biopsy, angiography, or systemic chemotherapy** are **not mistaken for TACE**.  
6. **Handle Uncertainty**: If the report lacks a direct mention of **TACE**, assume it **was not performed**.  
7. **Generate JSON Output**: Construct a response in strict JSON format, adhering to the following structure:  

\end{lstlisting}

\texttt{\normalsize Prompt for Question  22:}
\begin{lstlisting}[]

Your task is to determine whether **a transarterial radioembolization (TARE) procedure** was explicitly performed. **TARE is defined as the intra-arterial administration of radioactive microspheres, most commonly Yttrium-90 (Y90), to treat tumors.**  

#### **Key Criteria for TARE**  
- **Must involve intra-arterial administration of radioactive material** (e.g., Yttrium-90 microspheres).  
- **May include angiographic mapping with or without macroaggregated albumin (MAA) administration** to assess vascular anatomy before Y90 treatment.  
- **Typically performed for hepatic tumors (e.g., hepatocellular carcinoma, metastatic liver disease).**  

#### **Exclusion Criteria**  
- **Transarterial Chemoembolization (TACE)** - Uses **chemotherapy agents** instead of radioactive microspheres.  
- **Elective Embolization (e.g., uterine fibroid embolization, AVM embolization)** - Not the same as TARE.  
- **Emergency Embolization** - Performed for hemorrhage control, not tumor treatment.  

If the procedure was explicitly performed, return `1` along with an explanation. Otherwise, return `0` with an explanation.

Follow a structured reasoning approach to do the task:

### **Reasoning Steps**  
1. **Identify Mentions**: Extract all references to **TARE, transarterial radioembolization, Yttrium-90 (Y90) therapy, angiographic mapping, or MAA administration** in the report.  
2. **Confirm the Procedure Type**: Ensure the report explicitly states that **radioactive microspheres were administered intra-arterially** or that **angiographic mapping for TARE was performed.**  
3. **Differentiate from Other Procedures**:  
   - **Exclude TACE** (chemoembolization with doxorubicin, cisplatin, mitomycin C).  
   - **Exclude elective embolization for AVMs, fibroids, or varicoceles.**  
   - **Exclude emergency embolization for trauma or bleeding.**  
4. **Resolve Ambiguities**: If the report mentions **TARE was considered, planned, or recommended** but does not confirm that it was **performed**, classify it as **not performed**.  
5. **Exclude Non-Relevant Procedures**: Ensure that **angiography alone, hepatic artery catheterization, or systemic radiation therapy** are **not mistaken for TARE**.  
6. **Handle Uncertainty**: If the report is vague or does not explicitly mention of **TARE**, assume it **was not performed**.  
7. **Generate JSON Output**: Construct a response in strict JSON format, adhering to the following structure:  

\end{lstlisting}

\texttt{\normalsize Prompt for Question  23:}
\begin{lstlisting}[]

Your task is to determine whether an **intravascular procedure was performed** that **does not fit into** any of the following predefined categories:  

- **Venous access procedures** (e.g., port placement, exchange, or removal, central venous catheter procedures, dialysis catheter procedures).  
- **Dialysis access interventions** (e.g., fistulagrams, graft evaluations, declot procedures for arteriovenous access).  
- **IVC filter placement or removal**.  
- **Ablation procedures** (e.g., venous ablation).  
- **Stent or stent-graft placement** (arterial or venous).  
- **TIPS, DIPS, or other portosystemic shunt procedures**.  
- **Embolization procedures** (elective or emergency).  
- **Transarterial Chemoembolization (TACE)**.  
- **Transarterial Radioembolization (TARE)**.  

If a procedure **does not fall under any of these categories** and involves **intravascular access**, it qualifies for a **label of `1`**. Examples of qualifying procedures include:  
- **Transjugular liver or renal biopsy**  
- **Other intravascular biopsies**  
- **Fiducial marker placement using an intravascular approach**  

If the procedure was explicitly performed, return `1` along with an explanation. Otherwise, return `0` with an explanation.

Follow a structured reasoning approach to do the task:

### **Reasoning Steps**  
1. **Identify Any Intravascular Procedure**:  
   - Look for descriptions of procedures that involve catheter-based access to blood vessels.  

2. **Check for Exclusion Criteria**:  
   - If the procedure is **already classified** under venous access, dialysis interventions, IVC filters, ablations, stents, TIPS, embolization, TACE, or TARE, **do not count it**.  

3. **Check for Inclusion Criteria**:  
   - If the procedure is **not one of the excluded categories** but still involves **vascular access**, determine if it falls under examples such as **intravascular biopsy (e.g., transjugular liver biopsy, renal biopsy) or fiducial marker placement**.  

4. **Resolve Ambiguities**:  
   - If the report states that **a procedure was planned or considered** but does not confirm that it was **performed**, classify it as **not performed**.  

5. **Ensure No Misclassification**:  
   - Verify that **angiography, diagnostic imaging alone, or systemic procedures** are **not mistakenly included** as intravascular interventions.  

6. **Generate JSON Output**:  
   - Your response **must** be a **valid JSON object** with the following structure:  

\end{lstlisting}

\texttt{\normalsize Prompt for Question  24:}
\begin{lstlisting}[]

Your task is to determine whether a **percutaneous biopsy** was performed.  

- **Percutaneous biopsy** refers to a biopsy procedure where a needle is inserted **through the skin** to obtain a tissue sample from an organ or mass.  
- **Percutaneous biopsy does NOT include intravascular biopsy procedures** such as:  
  - **Transjugular liver biopsy**  
  - **Transjugular renal biopsy**  

If the procedure was explicitly performed, return `1` along with an explanation. Otherwise, return `0` with an explanation.

Follow a structured reasoning approach to do the task:

### **Reasoning Steps**  

1. **Identify Any Biopsy Procedure**:  
   - Look for key terms indicating a **biopsy** was performed. Common descriptors include:  
     - 'Fine-needle aspiration (FNA)'  
     - 'Core needle biopsy (CNB)'  
     - 'Image-guided biopsy' (e.g., ultrasound-guided, CT-guided)  

2. **Confirm Percutaneous Approach**:  
   - Ensure the biopsy was performed **through the skin** and not via a **vascular approach**.  
   - If terms like **'transjugular'**, **'intravascular'**, or **'endovascular'** appear, **do not count it** as percutaneous.  

3. **Resolve Ambiguities**:  
   - If the report states that a **biopsy was considered or planned** but does not confirm it was **performed**, classify it as **not performed**.  

4. **Ensure No Misclassification**:  
   - If the procedure involves **resection, excision, or surgical biopsy**, it is **not** a percutaneous biopsy and should be excluded.  
   - If only **biopsy results** are mentioned but **no procedural details** are given, assume the biopsy is **not documented** in this report.  

5. **Generate JSON Output**:  
   - Your response **must** be a **valid JSON object** with the following structure:  

\end{lstlisting}

\texttt{\normalsize Prompt for Question  25:}
\begin{lstlisting}[]

Your task is to determine whether an **abscess drainage procedure** was performed.  

- **Abscess drainage procedures** include:  
  - Placement of a drain for an **abscess** or **fluid collection**  
  - **Drain revision, repositioning, upsizing, or removal**  

- **Exclusions** (Do **not** count the following as abscess drainage):  
  - **Biliary drains and tubes**  
  - **Cholecystostomy drains and tubes**  
  - **Nephrostomy drains and tubes**  
  - **Nephroureterostomy drains and tubes**  
  - **Chest drains and tubes**  

If the procedure was explicitly performed, return `1` along with an explanation. Otherwise, return `0` with an explanation.

Follow a structured reasoning approach to do the task:

### **Reasoning Steps**  

1. **Identify Any Drainage Procedure**:  
   - Look for terms indicating **drain placement, removal, revision, or repositioning** related to an **abscess or fluid collection**.  
   - Common descriptors include:  
     - 'Percutaneous abscess drainage'  
     - 'Fluid collection drainage'  
     - 'Drain placed for abscess'  
     - 'Drain repositioning/upsize'  

2. **Confirm That It Is an Abscess Drainage Procedure**:  
   - Ensure the **drainage target** is an **abscess or fluid collection**, not an excluded category.  
   - If the report specifies a **biliary, cholecystostomy, nephrostomy, nephroureterostomy, or chest drain**, **do not count it** as abscess drainage.  

3. **Resolve Ambiguities**:  
   - If the report states that **drainage was planned or considered** but does **not confirm it was performed**, classify it as **not performed**.  
   - If **only fluid aspiration is performed without drain placement**, it should **not** be classified as a drainage procedure.  

4. **Ensure No Misclassification**:  
   - **Exclude surgical drainage procedures** if they were performed in an **operating room** instead of radiologically guided drainage.  
   - If the **report only mentions drain presence** without confirming **drain placement, revision, or removal**, assume it is **not documented** in this report.  

5. **Generate JSON Output**:  
   - Your response **must** be a **valid JSON object** with the following structure:  

\end{lstlisting}

\texttt{\normalsize Prompt for Question  26:}
\begin{lstlisting}[]

Your task is to determine whether a **paracentesis or thoracentesis** procedure was performed.  

- **Definitions**:  
  - **Paracentesis**: A procedure involving **needle drainage of peritoneal fluid (ascites)** from the **abdomen**.  
  - **Thoracentesis**: A procedure involving **needle drainage of pleural fluid (effusion)** from the **chest (pleural cavity)**.  

If the procedure was explicitly performed, return `1` along with an explanation. Otherwise, return `0` with an explanation.

Follow a structured reasoning approach to do the task:

### **Reasoning Steps**  

1. **Identify Any Drainage Procedure**:  
   - Look for terms indicating **fluid drainage** from the **abdomen (paracentesis)** or **pleural space (thoracentesis)**.  
   - Common descriptors include:  
     - 'Paracentesis performed'  
     - 'Thoracentesis completed'  
     - 'Fluid removed from the peritoneal cavity'  
     - 'Pleural effusion drained under ultrasound guidance'  

2. **Confirm That It Matches the Definition**:  
   - Ensure the report **explicitly states** that **paracentesis or thoracentesis was performed**.  
   - If the procedure is **planned or considered but not confirmed as performed**, classify it as **not performed**.  

3. **Exclude Unrelated Drainage Procedures**:  
   - **Do not classify as paracentesis/thoracentesis** if:  
     - **Fluid aspiration was performed without drainage**.  
     - The report describes **abscess drainage, biliary drainage, or nephrostomy**, which are **not** paracentesis/thoracentesis.  
     - A **chest tube** was placed instead of thoracentesis.  

4. **Resolve Ambiguities**:  
   - If the report **only mentions the presence of fluid (ascites or pleural effusion)** but does not confirm a **drainage procedure**, classify as **not documented**.  
   - If the **procedure attempt failed and no fluid was drained**, classify as **not performed**.  

5. **Generate JSON Output**:  
   - Your response **must** be a **valid JSON object** with the following structure:  

\end{lstlisting}

\texttt{\normalsize Prompt for Question  27:}
\begin{lstlisting}[]

Your task is to determine whether a **chest tube placement** procedure was performed.  

- **Definition**:  
  - **Chest tube placement** refers to the insertion of a tube into the **pleural space** to drain **air (pneumothorax), fluid (pleural effusion), blood (hemothorax), pus (empyema), or chyle (chylothorax)**.  
  - This is distinct from **thoracentesis**, which involves a needle aspiration without an indwelling tube.  

If the procedure was explicitly performed, return `1` along with an explanation. Otherwise, return `0` with an explanation.

Follow a structured reasoning approach to do the task:

### **Reasoning Steps**  

1. **Identify Any Tube Placement Procedure**:  
   - Look for terms indicating a **chest tube was inserted**.  
   - Common descriptors include:  
     - 'Chest tube placed'  
     - 'Thoracostomy tube inserted'  
     - 'Pleural drain inserted'  
     - 'Intercostal drain placement'  

2. **Confirm That It Matches the Definition**:  
   - Ensure that the **tube was actually placed into the pleural space**.  
   - **Exclude** procedures that only describe **thoracentesis (needle drainage without a tube)**.  

3. **Resolve Ambiguities**:  
   - If the report states that **a chest tube was planned or considered** but does **not confirm it was performed**, classify it as **not performed**.  
   - If the **report only mentions the presence of a chest tube** but does not confirm **placement in this report**, assume it is **not documented** in this instance.  

4. **Ensure No Misclassification**:  
   - **Do not classify as chest tube placement if:**  
     - The procedure was an **aspiration-only thoracentesis** without a tube.  
     - The report describes **pleural catheter placement**, which is distinct from a chest tube.  
     - The tube was placed **outside the pleural cavity**, such as an abdominal or mediastinal drain.  

5. **Generate JSON Output**:  
   - Your response **must** be a **valid JSON object** with the following structure:  

\end{lstlisting}

\texttt{\normalsize Prompt for Question  28:}
\begin{lstlisting}[]

Your task is to determine whether a **pleurodesis procedure** was performed.  

- **Definition**:  
  - **Pleurodesis** is a medical procedure that **induces adhesion between the pleural layers** to prevent the recurrence of pleural effusion or pneumothorax.  
  - This is typically achieved by **introducing a chemical agent (e.g., talc, doxycycline, or bleomycin) or mechanical abrasion into the pleural space**.  

If the procedure was explicitly performed, return `1` along with an explanation. Otherwise, return `0` with an explanation.

Follow a structured reasoning approach to do the task:
### **Reasoning Steps**  

1. **Identify Any Mention of Pleurodesis**:  
   - Look for terms indicating **pleurodesis was performed**.  
   - Common descriptors include:  
     - 'Chemical pleurodesis with talc/doxycycline performed'  
     - 'Mechanical pleurodesis via abrasion'  
     - 'Instillation of pleurodesis agent into pleural space'  

2. **Confirm That It Matches the Definition**:  
   - Ensure that the **procedure was actually carried out**, not just considered or planned.  
   - Verify that **the agent or mechanical technique used for pleurodesis is explicitly stated**.  

3. **Resolve Ambiguities**:  
   - If the report states that **pleurodesis was planned or considered** but does **not confirm it was performed**, classify it as **not performed**.  
   - If the **report only mentions a prior pleurodesis** without confirming that a **new pleurodesis was performed**, assume it is **not documented in this instance**.  

4. **Ensure No Misclassification**:  
   - **Do not classify as pleurodesis if:**  
     - The procedure **only describes chest tube placement** without mention of pleurodesis.  
     - The **pleural space was drained** (e.g., thoracentesis) but **no pleurodesis agent or abrasion was applied**.  
     - The procedure was a **pleural biopsy or another intervention unrelated to pleurodesis**.  

5. **Generate JSON Output**:  
   - Your response **must** be a **valid JSON object** with the following structure:  


\end{lstlisting}

\texttt{\normalsize Prompt for Question  29:}
\begin{lstlisting}[]

Your task is to determine whether a **procedure involving a biliary drain or biliary stent** was performed.  

- **Definition**:  
  - **Biliary drain or stent procedures** include:  
    - **Percutaneous transhepatic cholangiography (PTC)**  
    - **Biliary drain placement, evaluation, or revision**  
    - **Biliary stent placement or revision**  
  - **Exclusion**:  
    - **Cholecystostomy procedures (placement, exchange, or removal) do NOT count as biliary drain or stent procedures.**  

If the procedure was explicitly performed, return `1` along with an explanation. Otherwise, return `0` with an explanation.

Follow a structured reasoning approach to do the task:
### **Reasoning Steps**  

1. **Identify Mentions of a Biliary Drain or Stent Procedure**:  
   - Look for phrases such as:  
     - 'Percutaneous transhepatic cholangiography (PTC) performed'  
     - 'Biliary drain placement/revision/exchange'  
     - 'Biliary stent placement/revision'  
     - 'Intervention in the biliary system with drain or stent'  

2. **Confirm That It Matches the Definition**:  
   - Ensure that the procedure was **actually performed**, not just recommended or considered.  
   - Verify that **it involves the biliary system** and is **not related to the gallbladder (cholecystostomy)**.  

3. **Resolve Ambiguities**:  
   - If the report **mentions a prior biliary drain or stent** but does **not confirm a new procedure was performed**, classify as **not documented**.  
   - If the report **only discusses imaging of the biliary system (e.g., MRCP) without intervention**, classify as **not performed**.  

4. **Exclude Non-Biliary Procedures**:  
   - **Do not classify as a biliary drain or stent procedure if:**  
     - The procedure was a **cholecystostomy placement, exchange, or removal**.  
     - The report describes **only a diagnostic cholangiogram with no intervention**.  
     - The intervention was **related to another organ system (e.g., pancreatic duct stenting, nephrostomy placement, or GI stenting)**.  

5. **Generate JSON Output**:  
   - Your response **must** be a **valid JSON object** with the following structure:  

\end{lstlisting}

\texttt{\normalsize Prompt for Question  30:}
\begin{lstlisting}[]

Your task is to determine whether a **genitourinary drain procedure** was performed.  

- **Definition**:  
  - **Genitourinary drain procedures** include:  
    - **Nephrostomy tube placement, exchange, revision, or removal**  
    - **Nephroureteral stent placement, exchange, revision, or removal**  

If the procedure was explicitly performed, return `1` along with an explanation. Otherwise, return `0` with an explanation.

Follow a structured reasoning approach to do the task:

### **Reasoning Steps**  

1. **Identify Mentions of a Genitourinary Drain Procedure**:  
   - Look for phrases such as:  
     - 'Nephrostomy tube placed/exchanged/revised/removed'  
     - 'Nephroureteral stent placement/exchange/revision/removal'  
     - 'Percutaneous nephrostomy procedure performed'  
     - 'Intervention involving urinary drainage system'  

2. **Confirm That It Matches the Definition**:  
   - Ensure that the procedure was **actually performed**, not just recommended or considered.  
   - Verify that the procedure was **specific to the genitourinary system**, particularly **kidneys and ureters**.  

3. **Resolve Ambiguities**:  
   - If the report **mentions a prior nephrostomy tube or stent** but does **not confirm a new procedure**, classify as **not documented**.  
   - If the report **only discusses imaging of the genitourinary system (e.g., CT Urogram, MR Urogram) without intervention**, classify as **not performed**.  

4. **Exclude Non-Genitourinary Procedures**:  
   - **Do not classify as a genitourinary drain procedure if:**  
     - The procedure was **related to the bladder or urethra (e.g., Foley catheter placement, suprapubic catheter placement)**.  
     - The report describes **only diagnostic imaging without any interventional component**.  
     - The intervention was **related to another organ system (e.g., biliary drains, chest tubes, or gastrointestinal stents)**.  

5. **Generate JSON Output**:  
   - Your response **must** be a **valid JSON object** with the following structure:  

\end{lstlisting}

\texttt{\normalsize Prompt for Question  31:}
\begin{lstlisting}[]

Your task is to determine whether an **enterostomy procedure** was performed.  

- **Definition**:  
  - **Enterostomy procedures** include:  
    - **Gastrostomy placement, exchange, revision, or removal**  
    - **Gastrojejunostomy placement, exchange, revision, or removal**  
    - **Jejunostomy placement, exchange, revision, or removal**  
    - **Ileostomy placement, exchange, revision, or removal**  
    - **Cecostomy placement, exchange, revision, or removal**  

If the procedure was explicitly performed, return `1` along with an explanation. Otherwise, return `0` with an explanation.

Follow a structured reasoning approach to do the task:

### **Reasoning Steps**  

1. **Identify Mentions of an Enterostomy Procedure**:  
   - Look for phrases such as:  
     - 'Gastrostomy tube placed/exchanged/revised/removed'  
     - 'Gastrojejunostomy tube placement/exchange/revision/removal'  
     - 'Jejunostomy procedure performed'  
     - 'Percutaneous enterostomy tube placement'  
     - 'Ileostomy or cecostomy intervention performed'  

2. **Confirm That It Matches the Definition**:  
   - Ensure that the procedure was **actually performed**, not just recommended or considered.  
   - Verify that the procedure was **specific to the gastrointestinal system**, particularly the stomach, small intestine, or colon.  

3. **Resolve Ambiguities**:  
   - If the report **mentions a prior enterostomy procedure** but does **not confirm a new procedure**, classify as **not documented**.  
   - If the report **only discusses enteric imaging (e.g., fluoroscopy to check tube placement) without intervention**, classify as **not performed**.  

4. **Exclude Non-Enterostomy Procedures**:  
   - **Do not classify as an enterostomy procedure if:**  
     - The procedure was **related to the esophagus or rectum (e.g., esophageal stenting, rectal tube placement)**.  
     - The report describes **only diagnostic imaging without any interventional component**.  
     - The intervention was **related to another organ system (e.g., biliary or genitourinary procedures)**.  

5. **Generate JSON Output**:  
   - Your response **must** be a **valid JSON object** with the following structure:  

\end{lstlisting}

\texttt{\normalsize Prompt for Question  32:}
\begin{lstlisting}[]

Your task is to determine whether a **cholecystostomy procedure** was performed.  

- **Definition**:  
  - **Cholecystostomy procedures** include:  
    - **Cholecystostomy drain or tube placement**  
    - **Cholecystostomy drain or tube exchange**  
    - **Cholecystostomy drain or tube revision**  
    - **Cholecystostomy drain or tube removal**  
    - **Cholecystography (contrast imaging of the gallbladder through a cholecystostomy tube)**  

If the procedure was explicitly performed, return `1` along with an explanation. Otherwise, return `0` with an explanation.

Follow a structured reasoning approach to do the task:


### **Reasoning Steps**  

1. **Identify Mentions of a Cholecystostomy Procedure**:  
   - Look for phrases such as:  
     - 'Cholecystostomy tube placed/exchanged/revised/removed'  
     - 'Percutaneous cholecystostomy performed'  
     - 'Cholecystography performed via cholecystostomy tube'  

2. **Confirm That It Matches the Definition**:  
   - Ensure that the procedure was **actually performed**, not just recommended or considered.  
   - Verify that the procedure was **specific to the gallbladder**, involving intervention with a drain, tube, or contrast study.  

3. **Resolve Ambiguities**:  
   - If the report **mentions a prior cholecystostomy procedure** but does **not confirm a new procedure**, classify as **not documented**.  
   - If the report **only describes gallbladder imaging (e.g., ultrasound, CT, or MRI) without intervention**, classify as **not performed**.  

4. **Exclude Non-Cholecystostomy Procedures**:  
   - **Do not classify as a cholecystostomy procedure if:**  
     - The procedure was **related to the biliary system but did not involve a cholecystostomy drain or tube (e.g., biliary stenting, ERCP, PTC)**.  
     - The report describes **only gallbladder assessment without intervention**.  
     - The intervention was **related to another organ system (e.g., enterostomy or nephrostomy procedures)**.  

5. **Generate JSON Output**:  
   - Your response **must** be a **valid JSON object** with the following structure:  

\end{lstlisting}

\texttt{\normalsize Prompt for Question  33:}
\begin{lstlisting}[]

Your task is to determine whether a **cyst or lymphocele procedure** was performed.  

- **Definition**:  
  - **Cyst or lymphocele procedures** include:  
    - **Aspiration of a cyst or lymphocele**  
    - **Drain placement for a cyst or lymphocele**  
    - **Drain exchange, revision, or removal for a cyst or lymphocele**  
    - **Sclerosis of a cyst or lymphocele (injecting a sclerosing agent to obliterate the lesion)**  

If the procedure was explicitly performed, return `1` along with an explanation. Otherwise, return `0` with an explanation.

Follow a structured reasoning approach to do the task:

### **Reasoning Steps**  

1. **Identify Mentions of a Cyst or Lymphocele Procedure**:  
   - Look for phrases such as:  
     - 'Aspiration of cyst/lymphocele performed'  
     - 'Drain placed/exchanged/revised/removed for cyst/lymphocele'  
     - 'Sclerotherapy performed for cyst/lymphocele'  

2. **Confirm That It Matches the Definition**:  
   - Ensure that the procedure was **actually performed**, not just recommended or considered.  
   - Verify that the intervention was **specific to a cyst or lymphocele**.  

3. **Resolve Ambiguities**:  
   - If the report **mentions a prior cyst or lymphocele procedure** but does **not confirm a new procedure**, classify as **not documented**.  
   - If the report **only describes imaging (e.g., ultrasound, CT, or MRI) of a cyst or lymphocele without intervention**, classify as **not performed**.  

4. **Exclude Non-Cyst/Lymphocele Procedures**:  
   - **Do not classify as a cyst or lymphocele procedure if:**  
     - The procedure was **related to an abscess or other fluid collection (e.g., abscess drainage, seroma drainage, or pleural effusion drainage)**.  
     - The report describes **only diagnostic assessment without an interventional procedure**.  
     - The intervention was **related to another organ system (e.g., biliary drainage, peritoneal fluid drainage, nephrostomy, or cholecystostomy procedures)**.  

5. **Generate JSON Output**:  
   - Your response **must** be a **valid JSON object** with the following structure:  

\end{lstlisting}

\texttt{\normalsize Prompt for Question  34:}
\begin{lstlisting}[]

Your task is to determine whether a **nonvascular stent placement** was performed.  

- **Definition**:  
  - **Nonvascular stents** refer to stents placed in non-blood vessel structures, including:  
    - **Esophageal stent placement**  
    - **Tracheobronchial (airway) stent placement**  
    - **Duodenal stent placement**  
    - **Colonic stent placement**  

If the procedure was explicitly performed, return `1` along with an explanation. Otherwise, return `0` with an explanation.

Follow a structured reasoning approach to do the task:

### **Reasoning Steps**  

1. **Identify Mentions of a Nonvascular Stent Placement**:  
   - Look for terms such as:  
     - 'Esophageal stent placed'  
     - 'Tracheal/bronchial stent placement performed'  
     - 'Duodenal stent deployed'  
     - 'Colonic stent placement completed'  

2. **Confirm That It Matches the Definition**:  
   - Ensure that the procedure was **actually performed**, not just recommended or planned.  
   - Verify that the **stent was placed in a nonvascular structure**.  

3. **Resolve Ambiguities**:  
   - If the report **mentions a prior nonvascular stent placement** but does **not confirm a new procedure**, classify as **not documented**.  
   - If the report **only describes imaging (e.g., X-ray, fluoroscopy, or CT) of a stent without confirming a new placement**, classify as **not performed**.  

4. **Exclude Non-Stent and Vascular Stent Procedures**:  
   - **Do not classify as nonvascular stent placement if:**  
     - The procedure involves **vascular (arterial or venous) stents**, such as in coronary, carotid, iliac, or femoral arteries.  
     - The procedure involves **other non-stent interventions**, such as dilation without stent placement.  
     - The report describes **only a stent exchange, removal, or revision** without confirming a new stent placement.  

5. **Generate JSON Output**:  
   - Your response **must** be a **valid JSON object** with the following structure:  

\end{lstlisting}

\texttt{\normalsize Prompt for Question  35:}
\begin{lstlisting}[]

Your task is to determine whether a **miscellaneous nonvascular transplant intervention** was performed.  

- **Definition**:  
  - A **miscellaneous nonvascular transplant intervention** refers to **procedures performed exclusively on a transplant organ that do not involve arteries or veins**.  
  - The report **must explicitly confirm that the organ is a transplant organ** (not a native organ).  
  - **Examples include**:  
    - **Stent placement** involving a transplant organ  
    - **Balloon plasty procedures**, such as **ureteroplasty**  

If the procedure was explicitly performed, return `1` along with an explanation. Otherwise, return `0` with an explanation.

Follow a structured reasoning approach to do the task:
### **Reasoning Steps**  

1. **Identify Mentions of a Transplant Organ**:  
   - Look for explicit mentions that the **organ is a transplant**:  
     - Phrases like '**transplant kidney**,' '**transplant liver**,' or '**transplant ureter**' confirm that the organ is not native.  
   - If the report does **not** specify that the organ is a **transplant**, classify as **not documented (0)**.  

2. **Identify a Relevant Nonvascular Intervention**:  
   - Look for **nonvascular procedures** performed **on the transplant organ**:  
     - Stent placements in a transplant organ  
     - Balloon plasty procedures (e.g., **ureteroplasty**)  
   - **Do NOT include vascular procedures**, such as:  
     - **Transplant renal artery stenosis (TRAS) interventions**  
     - **Transjugular biopsy procedures**  
   - **Do NOT include** general drainage procedures such as:  
     - Nephrostomy placement/exchange/removal  
     - Nephroureterostomy placement/exchange/removal  
     - Biliary tube placement/exchange/removal  

3. **Confirm That the Procedure Was Performed**:  
   - The procedure must be **explicitly stated as completed**, not just planned or considered.  
   - If the report only describes **imaging or evaluation** without confirming an **intervention**, classify as **not performed**.  

4. **Resolve Ambiguities**:  
   - If the report **mentions a prior nonvascular transplant intervention** but does **not confirm a new procedure**, classify as **not documented**.  
   - If the procedure is **only vascular in nature**, classify as **not performed**.  

5. **Generate JSON Output**:  
   - Your response **must** be a **valid JSON object** with the following structure:  

\end{lstlisting}

\texttt{\normalsize Prompt for Question  36:}
\begin{lstlisting}[]

Your task is to determine whether a **tumor ablation** procedure was performed.  

- **Definition**:  
  - A **tumor ablation** is a procedure that destroys tumor tissue using thermal, chemical, or energy-based methods.  
  - **Examples of tumor ablation methods include**:  
    - **Radiofrequency ablation (RFA)**  
    - **Laser ablation**  
    - **Microwave ablation**  
    - **Cryoablation**  
    - **Ethanol injection/administration**  

If the procedure was explicitly performed, return `1` along with an explanation. Otherwise, return `0` with an explanation.

Follow a structured reasoning approach to do the task:
### **Reasoning Steps**  

1. **Identify Mentions of a Tumor Ablation Procedure**:  
   - Look for terms indicating that a **tumor ablation** was **performed**, such as:  
     - **'Radiofrequency ablation (RFA) performed'**  
     - **'Microwave ablation completed'**  
     - **'Cryoablation applied to tumor'**  
     - **'Ethanol injection administered'**  

2. **Differentiate from Other Procedures**:  
   - **Do NOT include** interventions that are **not ablations**, such as:  
     - **Surgical tumor resection**  
     - **Tumor embolization (TACE or TARE)**  
     - **Biopsy procedures**  
   - If the report only mentions **imaging, planning, or evaluation** for an ablation **without confirming its completion**, classify as **not performed**.  

3. **Confirm That the Procedure Was Performed**:  
   - The procedure must be **explicitly stated as completed**, not just planned or considered.  
   - If the report only describes an **intended** or **future** ablation, classify as **not performed**.  

4. **Resolve Ambiguities**:  
   - If the report mentions a **prior tumor ablation** but does not confirm a new procedure, classify as **not documented**.  
   - If the ablation is only referenced as a **potential treatment option** but was not performed, classify as **not performed**.  

5. **Generate JSON Output**:  
   - Your response **must** be a **valid JSON object** with the following structure:  

\end{lstlisting}

\texttt{\normalsize Prompt for Question  37:}
\begin{lstlisting}[]

Your task is to determine whether a **pain management procedure** was performed.  

- **Definition**:  
  - A **pain management procedure** is an interventional procedure aimed at alleviating pain.  
  - **Examples of pain management procedures include**:  
    - **Steroid injection**  
    - **Celiac plexus neurolysis**  
    - **Nerve cryoablation**  
    - **Periosteal cryoablation**  

If the procedure was explicitly performed, return `1` along with an explanation. Otherwise, return `0` with an explanation.

Follow a structured reasoning approach to do the task:

### **Reasoning Steps**  

1. **Identify Mentions of a Pain Management Procedure**:  
   - Look for terms indicating that a **pain management intervention** was **performed**, such as:  
     - **'Steroid injection administered'**  
     - **'Celiac plexus neurolysis performed'**  
     - **'Nerve cryoablation completed'**  
     - **'Periosteal cryoablation applied'**  

2. **Differentiate from Other Procedures**:  
   - **Do NOT include** interventions that are **not specifically pain management**, such as:  
     - **General anesthesia or sedation**  
     - **Surgical nerve decompression**  
     - **Tumor ablation procedures (unless explicitly performed for pain relief)**  
   - If the report only mentions **imaging, planning, or evaluation** for a pain management procedure **without confirming its completion**, classify as **not performed**.  

3. **Confirm That the Procedure Was Performed**:  
   - The procedure must be **explicitly stated as completed**, not just planned or considered.  
   - If the report only describes an **intended** or **future** pain management procedure, classify as **not performed**.  

4. **Resolve Ambiguities**:  
   - If the report mentions a **prior pain management procedure** but does not confirm a new one, classify as **not documented**.  
   - If the procedure is only referenced as a **potential treatment option** but was not performed, classify as **not performed**.  

5. **Generate JSON Output**:  
   - Your response **must** be a **valid JSON object** with the following structure:  

\end{lstlisting}

\texttt{\normalsize Prompt for Question  38:}
\begin{lstlisting}[]

Your task is to determine whether a **Fallopian tube recanalization** was performed.  

- **Definition**:  
  - **Fallopian tube recanalization (FTR)** is an interventional radiology procedure aimed at reopening blocked Fallopian tubes to restore fertility.  
  - **This procedure involves**:  
    - The use of **fluoroscopy and contrast dye** to visualize the Fallopian tubes.  
    - **Catheterization** or **balloon inflation** to clear obstructions.  

If the procedure was explicitly performed, return `1` along with an explanation. Otherwise, return `0` with an explanation.

Follow a structured reasoning approach to do the task:

### **Reasoning Steps**  

1. **Identify Mentions of a Fallopian Tube Recanalization Procedure**:  
   - Look for terms indicating that an **FTR procedure** was performed, such as:  
     - **'Fallopian tube recanalization completed'**  
     - **'Selective catheterization of the Fallopian tube performed'**  
     - **'Balloon dilation of Fallopian tube performed'**  

2. **Differentiate from Related Procedures**:  
   - **Do NOT include** procedures that are **not Fallopian tube recanalization**, such as:  
     - **Hysterosalpingography (HSG) alone** (without intervention)  
     - **Salpingectomy (Fallopian tube removal)**  
     - **In vitro fertilization (IVF) procedures**  
     - **General fertility evaluations or imaging**  

3. **Confirm That the Procedure Was Performed**:  
   - The procedure must be **explicitly stated as completed**, not just planned or considered.  
   - If the report only describes an **intended** or **future** Fallopian tube recanalization, classify as **not performed**.  

4. **Resolve Ambiguities**:  
   - If the report mentions a **prior Fallopian tube recanalization** but does not confirm a new one, classify as **not documented**.  
   - If the procedure is only referenced as a **potential treatment option** but was not performed, classify as **not performed**.  

5. **Generate JSON Output**:  
   - Your response **must** be a **valid JSON object** with the following structure:  

\end{lstlisting}

\texttt{\normalsize Prompt for Question  39:}
\begin{lstlisting}[]

Your task is to determine whether a **nonvascular, invasive procedure** was performed, while ensuring it **does not** fall into any of the **excluded categories** listed below.

- **Definition**:  
  - A **nonvascular, invasive procedure** involves the insertion of instruments or devices into the body, either through the skin or a natural orifice, **but does not** involve blood vessels (arteries or veins).  

- **Excluded Categories**:  
  - **Biopsy (percutaneous or otherwise)**
  - **Abscess drainage & tube checks**
  - **Paracentesis or thoracentesis**
  - **Chest tube placement or pleurodesis**
  - **Biliary interventions** (PTC, biliary drainage/stent procedures)
  - **Tube checks (for any previously placed drainage tubes)**
  - **Genitourinary interventions** (nephrostomy, nephroureteral tube placements/exchanges/removals)
  - **Gastrointestinal interventions** (gastrostomy, gastrojejunostomy, cholecystostomy)
  - **Cyst or lymphocele interventions**
  - **Nonvascular stents** (esophageal, tracheobronchial, duodenal, colonic)
  - **Transplant-related nonvascular interventions**
  - **Tumor ablation (RFA, microwave, cryoablation, ethanol)**
  - **Pain management procedures (nerve blocks, neurolysis, cryoablation)**
  - **Fallopian tube recanalization**
  - **Any intravascular procedure** (angioplasty, stenting, embolization, ablation, dialysis access, venous access, TIPS, DIPS, IVC filter procedures, TACE, TARE)

If the procedure was explicitly performed, return `1` along with an explanation. Otherwise, return `0` with an explanation.

Follow a structured reasoning approach to do the task:
### **Reasoning Steps**  

1. **Identify the Presence of an Invasive Procedure**:  
   - Look for descriptions of **instrumentation insertion** into the body through the skin or natural orifices.  
   - Common indicators include:  
     - **"Percutaneous intervention"**  
     - **"Endoscopic intervention"**  
     - **"Device placement"**  
     - **"Surgical instrument insertion"**  

2. **Check If the Procedure Falls Into an Excluded Category**:  
   - Compare the procedure against the **list of excluded interventions**.  
   - If the procedure **matches any exclusion**, classify as **not qualifying (return `0`)**.  

3. **Determine If the Procedure Is a Nonvascular, Invasive Procedure Not Listed as Excluded**:  
   - If it **does not** involve **arteries or veins** and is **not in the exclusion list**, classify as **qualifying (return `1`)**.  
   - Examples of qualifying procedures:  
     - **Foreign body removal from a nonvascular structure**  
     - **Fiducial marker placement (nonvascular target, such as lung or prostate)**  
     - **Percutaneous decompression of a non-abscess fluid collection (not cyst/lymphocele-related)**  

4. **Resolve Ambiguities**:  
   - If the report **only mentions a history of a procedure but no new procedure was performed**, classify as **not documented (`0`)**.  
   - If the procedure is **described as planned but not yet performed**, classify as **not performed (`0`)**.  

5. **Generate JSON Output**:  
   - Your response **must** be a **valid JSON object** with the following structure:  

\end{lstlisting}

\subsection{Example Report and Labels}
\texttt{Example Report}

\begin{lstlisting}[]
Pre-procedure diagnosis: Cholelithiasis
Post-procedure diagnosis: Same
Indication: Cholelithiasis
Additional clinical history: None

Complications: No immediate complications.

IMPRESSION:

*	Extensive cholelithiasis with partial removal of gallstones.
*	Stricture at the cystic duct, status post balloon dilation.
*	Appropriately positioned internal/external biliary drain.

Plan: 

*	Maintain biliary drain to gravity drainage. Flush twice daily to ensure tube patency.
*	Plan for staged gallstone lithotripsy and extraction.
PROCEDURE SUMMARY:
- Transhepatic cholangiogram via the existing access
- Cholangioscopy with gallstone lithotripsy and extraction. 
- Additional procedure(s): Internal/external biliary drain placement.

PROCEDURE DETAILS:

Pre-procedure
Consent: Informed consent for the procedure including risks, benefits and alternatives was obtained and time-out was performed prior to the procedure.
Preparation: The site was prepared and draped using maximal sterile barrier technique including cutaneous antisepsis.

Anesthesia/sedation
Level of anesthesia/sedation: General anesthesia
Anesthesia/sedation administered by: Anesthesiology
Total intra-service sedation time (minutes): See anesthesia flow sheet.

Cholangiogram, cholangioscopy, and gallstone lithotripsy/extraction.
Local anesthesia was administered. Contrast was injected through the indwelling biliary drain. A guidewire was passed through the gallbladder, cystic and common bile ducts, into the small bowel. Sequential dilation was performed, until ultimately a peel-away sheath was inserted into the gallbladder. A cholangioscope was advanced over the wire into the gallbladder. Electrohydraulic lithotripsy and basket retrieval of gallstones were performed. Due to extensive gallstone burden of large size and limited electrohydraulic lithotripsy devices, the visualized gallstones were incompletely extracted. A cholangiogram was subsequently performed through the peel-away sheath, demonstrating a focal area of stenosis at the cystic duct. Balloon dilation of the cystic and common duct was performed, with resolution of the previous area of stenosis.. A new internal/external biliary drainage catheter was placed, and the location of the distal tip was confirmed to be within small bowel following contrast injection.. 
Initial cholangiogram findings: Extensive gallstones burden. Contrast flows across the cystic duct into the duodenum.
New biliary drain: Cook internal/external biliary drain
New biliary drain diameter (French): 14
New biliary drain modifications: None
Final cholangiogram findings: Cholelithiasis. Internal/external biliary drain in place with contrast draining into the duodenum.
External catheter securement:  Non-absorbable suture

Contrast
Contrast agent: Omnipaque 300
Contrast volume (mL): 20

Radiation Dose
Fluoroscopy time (minutes): 17.0  
Reference air kerma (mGy): 185.1 
Kerma area product (uGy-m2): 2685.2

Additional Details
Additional description of procedure: None
Equipment details: None
Specimens removed: None
Estimated blood loss (mL): Less than 10
\end{lstlisting}


\texttt{Response from Claude-3.5-Haiku with CoT Prompting} \newline
\texttt{\normalsize Response for Question  1:}
\begin{lstlisting}[]
"No explicit mention or evidence of CTA or intra-operative cone-beam CT with contrast in the radiology report. Procedures were fluoroscopy-guided with contrast cholangiography."
\end{lstlisting}

\texttt{\normalsize Response for Question  2:}
\begin{lstlisting}[]
"No mention or evidence of MR Angiography (MRA) in the radiology report. The procedure involved endoscopic gallstone lithotripsy, cholangiography, and biliary drain placement using fluoroscopy."

\end{lstlisting}

\texttt{\normalsize Response for Question  3:}
\begin{lstlisting}[]
"No noninvasive vascular lab study (duplex ultrasound, color flow study, or PVR) was mentioned or performed in this radiology report, which focused on an endoscopic gallstone procedure."

\end{lstlisting}

\texttt{\normalsize Response for Question  4:}
\begin{lstlisting}[]
"No cardiac imaging study was performed in this radiology report, which details an endoscopic gallstone lithotripsy and biliary intervention procedure."

\end{lstlisting}

\texttt{\normalsize Response for Question  5:}
\begin{lstlisting}[]
"No arteriography was performed. The procedure was an endoscopic gallstone lithotripsy with biliary cholangiography, using contrast in the biliary system, not an arterial study."

\end{lstlisting}

\texttt{\normalsize Response for Question  6:}
\begin{lstlisting}[]
"No venography was performed. The procedure was an endoscopic gallstone lithotripsy with cholangiography using biliary contrast, not a venous imaging study."

\end{lstlisting}

\texttt{\normalsize Response for Question  7:}
\begin{lstlisting}[]
"No dialysis access evaluation was performed in this radiology report, which describes a biliary procedure involving gallstone lithotripsy and drain placement."

\end{lstlisting}

\texttt{\normalsize Response for Question  8:}
\begin{lstlisting}[]
"No carotid artery imaging was performed or mentioned in this radiology report, which describes an endoscopic gallstone lithotripsy procedure with biliary interventions."

\end{lstlisting}

\texttt{\normalsize Response for Question  9:}
\begin{lstlisting}[]
"No central venous catheter procedure was performed. The report details a biliary drain procedure with cholangioscopy and gallstone lithotripsy."
\end{lstlisting}

\texttt{\normalsize Response for Question  10:}
\begin{lstlisting}[]
"No IVC filter placement or removal was mentioned in the radiology report. The procedure focused on gallstone lithotripsy and biliary drain placement."

\end{lstlisting}

\texttt{\normalsize Response for Question  11:}
\begin{lstlisting}[]
"The radiology report describes a biliary intervention for gallstone management, with no mention of any venous ablation procedure"

\end{lstlisting}

\texttt{\normalsize Response for Question  12:}
\begin{lstlisting}[]
"No dialysis access intervention was performed. This report describes a biliary procedure involving gallstone lithotripsy and biliary drain placement."

\end{lstlisting}

\texttt{\normalsize Response for Question  13:}
\begin{lstlisting}[]
"No portosystemic shunt intervention was performed. The procedure was a biliary intervention involving gallstone lithotripsy and biliary drain placement."

\end{lstlisting}

\texttt{\normalsize Response for Question  14:}
\begin{lstlisting}[]
"The report describes an endoscopic gallstone lithotripsy with biliary drain placement and balloon dilation of biliary ducts, but no arterial angioplasty or stent placement was performed."

\end{lstlisting}

\texttt{\normalsize Response for Question  15:}
\begin{lstlisting}[]
"The radiology report describes a biliary intervention with no venous angioplasty or stent placement performed"

\end{lstlisting}

\texttt{\normalsize Response for Question  16:}
\begin{lstlisting}[]
"The radiology report describes an endoscopic gallstone lithotripsy and biliary drain placement procedure, with no mention of carotid artery stenting."

\end{lstlisting}

\texttt{\normalsize Response for Question  17:}
\begin{lstlisting}[]
"No thrombolytic therapy or thrombectomy was performed in this biliary intervention procedure focused on gallstone management"

\end{lstlisting}

\texttt{\normalsize Response for Question  18:}
\begin{lstlisting}[]
"No evidence of aortic endograft placement or revision in this radiology report. The procedure was a biliary intervention involving gallstone lithotripsy and drain placement."

\end{lstlisting}

\texttt{\normalsize Response for Question  19:}
\begin{lstlisting}[]
"No emergency embolization was performed. The procedure was a gallstone lithotripsy with biliary drain placement."

\end{lstlisting}

\texttt{\normalsize Response for Question  20:}
\begin{lstlisting}[]
"No evidence of elective embolization in this radiology report. The procedure was a gallstone lithotripsy and biliary intervention."
\end{lstlisting}

\texttt{\normalsize Response for Question  21:}
\begin{lstlisting}[]
"The report describes an endoscopic gallstone lithotripsy and biliary drain procedure. There is no mention of TACE, chemotherapy administration, or tumor embolization. The procedure focuses on gallstone removal and biliary drainage, which is entirely unrelated to transarterial chemoembolization."

\end{lstlisting}

\texttt{\normalsize Response for Question  22:}
\begin{lstlisting}[]
"This report describes an endoscopic gallstone lithotripsy with biliary drain placement. No evidence of transarterial radioembolization (TARE) was present."

\end{lstlisting}

\texttt{\normalsize Response for Question  23:}
\begin{lstlisting}[]
"This is a biliary procedure involving cholangioscopy and gallstone lithotripsy, with no intravascular intervention performed"

\end{lstlisting}

\texttt{\normalsize Response for Question  24:}
\begin{lstlisting}[]
"No percutaneous biopsy was performed in this radiology report. The procedure was focused on gallstone lithotripsy, cholangioscopy, and biliary drain placement."

\end{lstlisting}

\texttt{\normalsize Response for Question  25:}
\begin{lstlisting}[]
"This is a biliary drain placement procedure, which is explicitly excluded from abscess drainage procedures according to the defined criteria."

\end{lstlisting}

\texttt{\normalsize Response for Question  26:}
\begin{lstlisting}[]
"No paracentesis or thoracentesis was performed. The report describes an endoscopic gallstone lithotripsy with biliary drain placement, which is a different type of procedure."

\end{lstlisting}

\texttt{\normalsize Response for Question  27:}
\begin{lstlisting}[]
"This radiology report describes a biliary drain placement procedure, not a chest tube placement. The tube was placed in the biliary system, not the pleural space."

\end{lstlisting}

\texttt{\normalsize Response for Question  28:}
\begin{lstlisting}[]
"No pleurodesis procedure was performed. The report describes an endoscopic gallstone lithotripsy and biliary drain placement, with no mention of any pleural space intervention."

\end{lstlisting}

\texttt{\normalsize Response for Question  29:}
\begin{lstlisting}[]
"An internal/external biliary drain was explicitly placed during the procedure, with detailed documentation of drain type, size, and placement confirmation"

\end{lstlisting}

\texttt{\normalsize Response for Question  30:}
\begin{lstlisting}[]
"The report describes a biliary intervention with biliary drain placement, which is not a genitourinary drain procedure. No nephrostomy tubes or nephroureteral stents were placed."

\end{lstlisting}

\texttt{\normalsize Response for Question  31:}
\begin{lstlisting}[]
"The report describes a biliary intervention with cholangioscopy and drain placement, but no enterostomy procedure was performed."

\end{lstlisting}

\texttt{\normalsize Response for Question  32:}
\begin{lstlisting}[]
"While a biliary drain was placed, this was not a specific cholecystostomy procedure. The intervention involved cholangioscopy, gallstone lithotripsy, and biliary drain placement, but did not meet the strict definition of a cholecystostomy."

\end{lstlisting}

\texttt{\normalsize Response for Question  33:}
\begin{lstlisting}[]
"This report describes a biliary intervention for gallstones, with no cyst or lymphocele procedure performed. The procedure involved cholangioscopy, gallstone lithotripsy, and biliary drain placement."

\end{lstlisting}

\texttt{\normalsize Response for Question  34:}
\begin{lstlisting}[]
"An internal/external biliary drain (nonvascular stent) was placed during the procedure, specifically a Cook 14 French biliary drain"

\end{lstlisting}

\texttt{\normalsize Response for Question  35:}
\begin{lstlisting}[]
"No transplant organ was mentioned; this was a procedure on a native biliary system"

\end{lstlisting}

\texttt{\normalsize Response for Question  36:} \textcolor{red}{\textbf{False Positive}}
\begin{lstlisting}[]
"No tumor ablation procedure was performed. The report details an endoscopic gallstone lithotripsy and biliary drain placement procedure."
\end{lstlisting}

\texttt{\normalsize Response for Question  37:}
\begin{lstlisting}[]
"No pain management procedure was performed. The report details a gallstone lithotripsy and biliary intervention without any specific pain management intervention."
\end{lstlisting}

\texttt{\normalsize Response for Question  38:}
\begin{lstlisting}[]
"The radiology report describes a biliary procedure involving gallstone lithotripsy and does not mention any Fallopian tube intervention."
\end{lstlisting}

\texttt{\normalsize Response for Question  39:}
\textcolor{red}{\textbf{False Negative}}
\begin{lstlisting}[]
"The procedure involves biliary interventions, specifically biliary drainage and drain placement, which are explicitly excluded from qualifying as a nonvascular, invasive procedure."
\end{lstlisting}

\end{document}